\documentclass{article} 
\usepackage{iclr2024_conference,times}
\usepackage{epsfig}
\usepackage{graphicx}
\usepackage{amsmath}
\usepackage{amssymb}
\usepackage{multirow}
\usepackage{graphics}
\usepackage{threeparttable}
\usepackage{color}
\usepackage[normalem]{ulem}
\usepackage{float}
\usepackage{amsfonts}
\usepackage{bm}
\usepackage{bbm}
\usepackage{enumitem}
\usepackage{algorithmic,algorithm}
\usepackage{natbib}
\usepackage{subcaption}
\usepackage{algorithm}
\usepackage{array}
\usepackage{colortbl}
\usepackage{pifont}
\usepackage{booktabs}
\usepackage{mathtools}
\usepackage{siunitx}
\usepackage{caption} 
\usepackage{xcolor}
\usepackage{subcaption} 
\usepackage[nopar]{lipsum}
\usepackage{listings}
\usepackage[export]{adjustbox}
\usepackage{makecell}

\usepackage{mathtools}
\usepackage{siunitx}
\usepackage[nopar]{lipsum}
\usepackage{listings}

\usepackage{amsmath,amsfonts,bm}

\def\eqref#1{equation~\ref{#1}}

\def\1{\bm{1}}

\DeclareMathAlphabet{\mathsfit}{\encodingdefault}{\sfdefault}{m}{sl}
\SetMathAlphabet{\mathsfit}{bold}{\encodingdefault}{\sfdefault}{bx}{n}

\usepackage{hyperref}
\usepackage{url}
\def\Model{UniDG }
\usepackage{xcolor}
\usepackage{wrapfig} 
\definecolor{myy}{RGB}{126,95,0}
\definecolor{mygray}{gray}{.9}
\definecolor{Gray}{gray}{0.9}
\definecolor{bblue}{RGB}{30,80,120}
\definecolor{mygray1}{gray}{.7}
\definecolor{ggray}{RGB}{127,127,127}
\definecolor{defaultcolor}{gray}{.9}
\definecolor{dark-gray}{gray}{0.20}
\newcommand{\pub}[1]{{\color{dark-gray}{\tiny{[{#1}]}}}}

\newcommand{\reshl}[2]{
	\textbf{#1} \fontsize{7.5pt}{1em}\selectfont\color{mygreen}{$\uparrow$ \textbf{#2}}
}
\definecolor{mygreen}{HTML}{39b54a}

\newcolumntype{x}[1]{>{\centering\arraybackslash}p{#1pt}}
\newcolumntype{y}[1]{>{\raggedright\arraybackslash}p{#1pt}}
\newcolumntype{z}[1]{>{\raggedleft\arraybackslash}p{#1pt}}

\newlength\savewidth

\title{Towards Unified and Effective Domain Generalization}

\author{
Yiyuan Zhang\textsuperscript{1,2}\thanks{Equal Contribution},
~~~ Kaixiong Gong\textsuperscript{1,2}$^\ast$,
~~~ Xiaohan Ding\textsuperscript{4},\\
~~\textbf{Kaipeng Zhang}\textsuperscript{2}\thanks{Correspondence},
~~~\textbf{Fangrui Lv}\textsuperscript{5},
\quad~~\textbf{Kurt Keutzer}\textsuperscript{3},
\quad~~ \textbf{Xiangyu Yue}\textsuperscript{1$\dag$}\\
\textsuperscript{1}Multimedia Lab, The Chinese University of Hong Kong\\
\textsuperscript{2}OpenGVLab, Shanghai AI Lab~~~
\textsuperscript{3}UC Berkeley
\textsuperscript{4}Tencent AI Lab
\textsuperscript{5}Tsinghua University\\
{\tt\small \{yiyuanzhang.ai, kaixionggong\}@gmail.com}
~~~{\tt\small xyyue@ie.cuhk.edu.hk}\\
\quad \quad ~~~\url{https://invictus717.github.io/Generalization}
}
\iclrfinalcopy 
\begin{document}

	\maketitle
	
	\begin{abstract}
        We propose \textbf{UniDG}, a novel and \textbf{Uni}fied framework for \textbf{D}omain \textbf{G}eneralization that is capable of significantly enhancing the out-of-distribution generalization performance of foundation models regardless of their architectures. The core idea of UniDG is to finetune models during the inference stage, which saves the cost of iterative training. Specifically, we encourage models to learn the distribution of test data in an unsupervised manner and impose a penalty regarding the updating step of model parameters. The penalty term can effectively reduce the catastrophic forgetting issue as we would like to maximally preserve the valuable knowledge in the original model. Empirically, across 12 visual backbones, including CNN-, MLP-, and Transformer-based models, ranging from 1.89M to 303M parameters, UniDG shows an average accuracy improvement of +5.4\% on DomainBed. These performance results demonstrate the superiority and versatility of UniDG. The code is publicly available at \url{https://github.com/invictus717/UniDG}.
	\end{abstract}
	
	\section{Introduction} ~\label{sec:intro}
    The Out-Of-Distribution (OOD) problem is a prevalent topic in the machine learning and computer vision communities~\citep{long2015learning,saito2020universal,sun2016deep,ebrahimi2020adversarial} as models of various architectures and scales are suffering from this problem~\citep{zhou2022domain,li2022sparse,chen2022compound,peng2022semantic}. Therefore, training deep models to generalize well on new domains has become a prevalent research topic~\citep{long2015learning,li2018domain,wang2019transferable,chen2022contrastive,cha2021swad,cha2022domain}. 
    \begin{wrapfigure}{r}[0cm]{0pt}
    \begin{minipage}{0.49\linewidth}
		\vspace{-1mm}
        \includegraphics[width=1.0\linewidth]{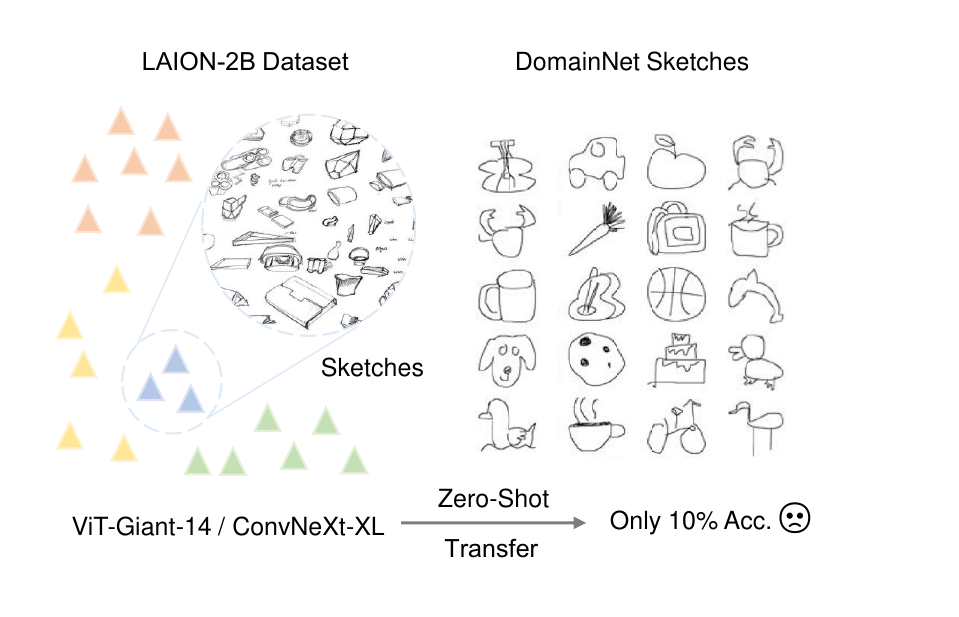}
        \vspace{-5.5mm}
        \caption{Large-scale pretrained foundation models are still suffering from domain shifts.}
        \label{fig:moti}
        \vspace{-5mm}
    \end{minipage}
    \end{wrapfigure}
    To overcome the domain shift problem, pretraining-based methods~\citep{radford2021learning,singh2022revisiting,cha2022domain} utilize large-scale data to obtain better generalization ability. However, in practice, domain shift can be so significant that even though the powerful foundation models have been pretrained on huge-scale datasets, directly generalizing the models to new domains still delivers unsatisfactory performance, as shown in Figure~\ref{fig:moti}.
    Another drawback of pretraining-based methods is the inferior finetuning performance, finetuning pretrained models leads to catastrophic forgetting and limited improvement on new domains~\citep{cha2022domain,li2022uncertainty,chen2022self}. As a workaround, pretraining-based methods may add data from the new domains into the pretraining dataset and retrain the models from scratch~\citep{shu2023clipood}. When the pretraining dataset is large (\textit{e.g.}, CLIP~\citep{radford2021learning} uses LAION-400M), this approach becomes significantly expensive.
    
	In contrast to the pretraining-based methods, Test-Time Adaptation (TTA) ~\citep{sun2020test,wang2020tent,wang2022continual,wang2022generalizing} is an alternative to mitigate domain shift on new domains. First, TTA requires no pretraining with novel data, and can directly leverage the off-the-shelf models. Second, by updating parameters in both training and evaluation stages~\citep{sun2020test}, TTA reduces the reliance of models on annotations in new domains. However, we would like to note several drawbacks of existing TTA methods.
    \begin{wrapfigure}{l}[0cm]{0pt}
    \begin{minipage}{0.43\linewidth}
    \vspace{-5.5mm}
    \includegraphics[width=1.0\linewidth]{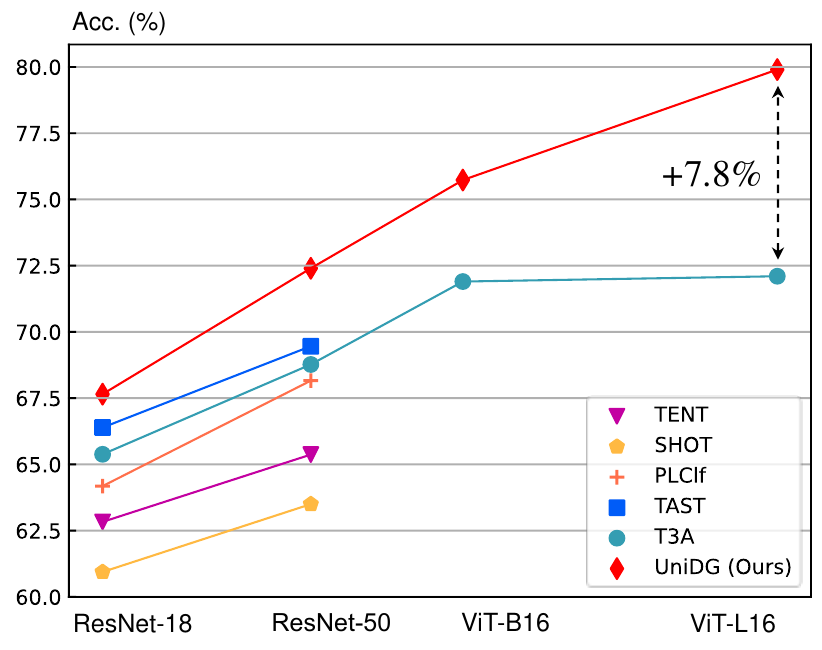}
    \vspace{-6mm}
    \caption{A comparison between existing methods and UniDG on the accuracy averaged across the PACS, VLCS, OfficeHome, and TerraInc datasets.}
    \label{fig:cmp}
    \vspace{-4mm}
    \end{minipage}
    \end{wrapfigure}
     \textbf{1)} Most TTA methods~\citep{wang2020tent,iwasawa2021test,jang2022test} require updating Batch Normalization (BN) \citep{ioffe2015batch} layers in the original model to adapt to the distribution of test data. However, recent visual foundation models such as vision transformers~\citep{dosovitskiy2020image} are developed with Layer Normalization (LN) layers. Due to the essential difference between BN and LN, simply adapting the ideas of BN-based methods to LN layers results in minimal improvement (around 0.5\%, see Appendix~\S~\ref{sec:supp:exp}). \textbf{2)} Recent TTA methods~\citep{zhang2023adanpc,park2023test,zhang2023domainadaptor, chen2023improved} show limited scalability on common visual foundation models ranging from small to large scales.
     For example, with TTA, only limited improvements (less than 2\%) on large-scale foundation models~\citep{radford2021learning,liu2022convnet} are observed. \textbf{3)} From a theoretical perspective, we find these TTA methods reduce the Neural Tangent Kernel~\citep{jacot2018neural} in the adaptation process, which limits the further generalization (theoretical analysis is presented in Appendix~\S~\ref{sec:theory}).

    \begin{wrapfigure}{l}[0cm]{0pt}
    \begin{minipage}{0.48\linewidth}
    \vspace{-5.5mm}
    \includegraphics[width=1.0\linewidth]{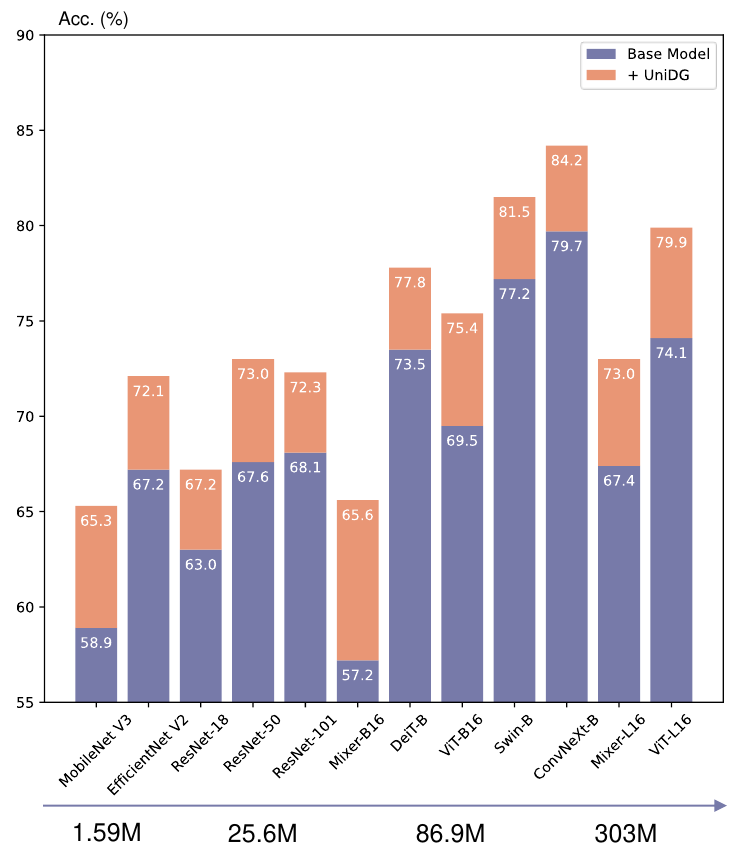}
    \vspace{-5.5mm}
    \caption{UniDG brings out an average of 5.4\% improvement to 12 backbones which scale from 1.59M to 303M parameters.}
    \label{fig:backbone}
    \vspace{-5mm}
    \end{minipage}
    \end{wrapfigure}
 To address the aforementioned drawbacks, we focus on an important topic of TTA method - how to effectively update the encoder (\textit{i.e.}, feature extractor) for TTA. Prior works either update the encoder via back-propagation or freeze it, but either way has its own weaknesses. 1) If we allow the encoder to update, similar to the weakness of finetuning a pretrained encoder as discussed above, catastrophic forgetting can happen during TTA and result in a significantly lower quality of extracted features. 2) When the encoder is frozen, it struggles to adapt effectively to new domains. Consequently, the extracted features require further refinement through additional mechanisms to be optimally utilized by the classifier.

 In this paper, we propose a novel method, named \textbf{Marginal Generalization}, to update the encoder for TTA. Intuitively, Marginal Generalization aims to let the encoder learn representations of the target data \emph{within a certain distance from the representations obtained by the initial model}. Here we use a simplified notation for brevity. Let $\sigma$ be the specified distance, $f(\cdot)$ be the fixed initial encoder and $f^\prime(\cdot)$ be a learnable copy of $f(\cdot)$, $x$ be the samples of the target domain, $q(\cdot)$ be the classifier which takes the representations $f^\prime(x)$ as inputs, the objective is to
    \begin{equation}\label{eq-intro}
        \begin{aligned}
            &\text{minimize}\quad [\text{entropy}(\text{softmax}(q(f^\prime(x))))]  &s.t.\hspace{1.5mm} \| f^\prime(x) - f(x)\|_F \leq \sigma \,.
        \end{aligned}
    \end{equation}

By doing so, we overcome the drawbacks of the two traditional approaches mentioned above. 
1) Intuitively, while the encoder $f^\prime(\cdot)$ is trying to adapt to the novel data, it always refers to the original model $f(\cdot)$ and keeps the representations within a distance $\sigma$ from the original, which means the pretrained source knowledge can be preserved and catastrophic forgetting is avoided. 2) As we keep updating the encoder via entropy minimization on the test data, it cooperates better with the classifier and yields more discriminative features on the target domain.

We would like to note that Marginal Generalization is \textbf{uni}versal because it does not require any specific structures in the original model nor the properties of the data, as well as effective (achieving improvement of 3.3\% on average accuracy as shown in Table~\ref{tab:ablation}). In addition, the features extracted by the updated encoder can be utilized by multiple TTA mechanisms. For example, by naturally combining Marginal Generalization and Memory Bank~\citep{wu2018unsupervised}, we propose \textbf{Differentiable Memory Bank}, which demonstrates superior performance over the traditional memory bank methods. For example, compared with T3A~\citep{iwasawa2021test} that adopts typical memory bank, our method with ResNet-50 backbone outperforms it by 4.3\% on average accuracy across 4 datasets as shown in Table~\ref{tab:tta}. Intuitively, UniDG simultaneously utilizes i) the local minimum of distances between adapted and source representations, and ii) the local maximum of information entropy between adapted representations and pseudo labels in the test-time process to continuously approximate the models on the test data with preserved pretrained knowledge. 
The details will be presented in Section~\ref{sec:method:feat}, and \ref{sec:method:dyn}.

Based on Marginal Generalization, we propose a framework composed of an adaptation method of the encoder (which is a \textbf{uni}versal method to extract better features) and Differentiable Memory Bank (which is a \textbf{uni}versal mechanism to refine features for DG) so that the framework is named \textbf{UniDG}, which delivers state-of-the-art performance on multiple domain generalization benchmarks. For example, UniDG delivers an average accuracy of 79.6\% on 5 widely adopted benchmarks including PACS, VLCS, and OfficeHome, outperforming the second-best CAR-FT~\citep{mao2022context} by 1.0\%. Additionally, UniDG is an architecture-agnostic framework that consistently yields significant improvements when applied to a wide range of visual backbones, including models of varying scales such as MobileNet V3~\citep{howard2019searching}, ConvNeXt-Base~\citep{liu2022convnet}, and ViT-Large~\citep{dosovitskiy2020image}, demonstrating its strong scalability as shown in Figure~\ref{fig:backbone}. For example, UniDG improves the mean accuracy scores by 5.4\% with such 12 models on PACS~\citep{torralba2011unbiased}, VLCS~\citep{li2017deeper}, OfficeHome~\citep{venkateswara2017deep}, and TerraInc~\citep{beery2018recognition} as shown in Figure~\ref{fig:cmp}. We would like to note that Marginal Generalization and Differentiable Memory Bank can also be used separately and combined with other methods. When we combine these two schemes, we observe an average improvement of +5.0\%, as shown in Table~\ref{tab:dg}.
   
    Our contributions are summarized as the following:
    \begin{itemize}
        \item We propose Marginal Generalization, which significantly mitigates the problems of feature encoder adaptation during TTA.
        \item With Marginal Generalization, we naturally upgrade the traditional memory bank mechanism to Differentiable Memory Bank and propose a universal TTA framework named UniDG.
        \item UniDG consistently outperforms the previous state-of-the-art methods by a significant margin (\textit{e.g.} +5.4\% on DomainBed), and it is applicable to a wide range of models with different architectures and varying scales.
        \item We show that the components in UniDG can also be separately combined with other methods (\textit{e.g.} averagely bringing out +5.0\% on DomainBed), demonstrating its flexibility.
    \end{itemize}

    \section{Method} ~\label{sec:method}
    
\begin{figure}[t]
        \centering
        \includegraphics[width=1.0\linewidth]{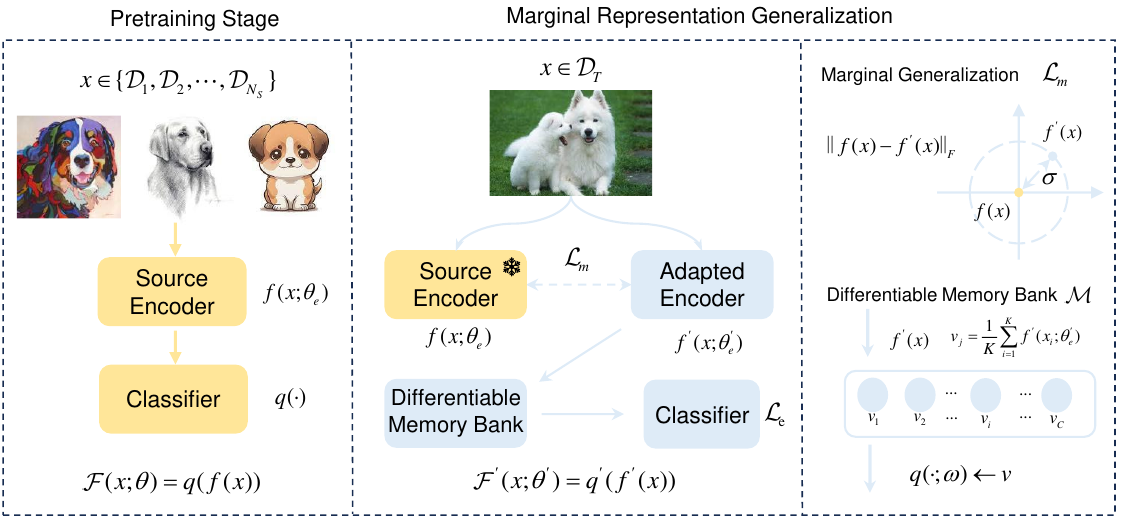}
    \caption{Illustration of UniDG, which consists of Marginal Generalization (\S~\ref{sec:method:feat}) and the Differentiable Memory Bank mechanism (\S~\ref{sec:method:dyn}).}
        \label{fig:framework}
        \vspace{-4mm}
    \end{figure}

    We first introduce the formulation of domain generalization and test-time adaptation in \S~\ref{sec:method:pre}. The framework of UniDG comprises two components: 1) we employ \emph{Marginal  Generalization} (\S~\ref{sec:method:feat}) to adapt the encoder, 2) we utilize prototypes with \emph{Differentiable Memory Bank } (\S~\ref{sec:method:dyn}) for learning a discriminative classifier on the target domain. The whole framework is summarized in Figure~\ref{fig:framework}. 
    
    \subsection{Preliminary} \label{sec:method:pre}
    
    \paragraph{Domain Generalization.}
    Given a set of source domains $ \mathcal{D}_S = \{\mathcal{D}_1, \mathcal{D}_2, \cdots, \mathcal{D}_{N} \}$, each domain $\mathcal{D}_j$ containing images and labels, $\{(\bm{x}_i, y_i)\}_{i=1}^{\|\mathcal{D}_j\|}$, where $\bm{x}_i$ denotes an image and $y_i$ indicates the corresponding ground truth label, the goal of DG is to train a model on source domains $\mathcal{D}_S$ in a way that it can effectively generalize to a novel target domain $\mathcal{D}_T$ which differs from any of the source domains. We denote the mapping function of the model as $\mathcal{F}: \bm{x}\rightarrow \bm{p} \in \mathbb{R}^{C}$, where $\bm{p}$ is the prediction and $C$ is the number of categories. $\mathcal{F}$ consists of two steps: feature extraction with the encoder $f(\cdot)$ and prediction with the classifier $q(\cdot)$ based on the features. Let $\theta$ be the parameters, $\mathcal{F}$ can be formulated as $\mathcal{F}(\bm{x};\theta) = q(f(\bm{x}))$. 

    \paragraph{Training on source domains.}
    Use $\ell_{\text{CE}}(\cdot)$ to denote the cross-entropy function, and the objective of training on the source domains is to optimize $\theta$ as
    \begin{equation}
    	\theta^{*} = \mathop{\arg \min}_{\theta} \mathbb{E}_{(\bm{x},y)\in \mathcal{D}_S}[\ell_{\text{CE}}(\mathcal{F}(\bm{x};\theta),y)] \,.
    	\label{eq:train_source}
    \end{equation}

    \paragraph{Test-Time Adaptation.}
   With $\theta^{*}$ trained on the source domains $\mathcal{D}_S$, test-time adaptation is a self-supervised learning process to further adapt parameters to the target domain $\mathcal{D}_T$. The encoder parameters during test time can be optimized as the following, where $\ell_{\text{TTA}}(\cdot)$ is the softmax entropy:
    \begin{equation}
        \begin{aligned}
                &\theta^\prime = \mathop{\arg \min}_{\theta} \mathbb{E}_{(\bm{x})\in \mathcal{D}_T}[\ell_{\text{TTA}}(\mathcal{F}^\prime(\bm{x}; \theta))] \,.
            \end{aligned}
        \label{eq:tta_target}
    \end{equation}

    \subsection{Marginal Generalization}~\label{sec:method:feat}
    Marginal Generalization aims to constrain the discrepancy between features extracted by the source encoder $f$ and the adapted encoder $f^\prime$ during the adaptation process so that the adapted model will be able to maintain general representation bias and relieve catastrophic forgetting while updating parameters. Here we adopt Euclidean distance as the metric out of its simplicity and universality, which is formulated with the Frobenius norm $\|\cdot\|_F$. We use $\theta_e$ to denote the parameters of the encoder, which is a subset of $\theta$, so that the encoder can be formulated as $f(\cdot;\theta_e)$. Given the pre-defined distance threshold $\sigma$, the objective then becomes
    \begin{equation}
        \begin{aligned}
                &\theta^\prime = \mathop{\arg \min}_{\theta} \mathbb{E}_{(\bm{x})\in \mathcal{D}_T}[\ell_{\text{TTA}}(\mathcal{F}^\prime(\bm{x}; \theta))] &s.t.\hspace{1.5mm} \| f^\prime(x;\theta_e^\prime) - f(x;\theta_e)\|_F \leq \sigma \,.
            \end{aligned}
        \label{eq:para_target}
    \end{equation}
    The motivation is that we desire to gradually update the parameters of the adapted encoder under the condition that the representation bias will not get sharply adapted. For the source feature extractor $f(\cdot;\theta_e)$, we freeze it and still use it to extract the representation from target domains as pretrained knowledge. For the adapted encoder $f^\prime(\cdot;\theta_e^\prime)$, we initialize it with the source-pretrained parameters $\theta_e$. Therefore, the discrepancy between the original and adapted representations can be formulated as the distance between $f(x;\theta_e)$ and $f^\prime(x;\theta_e^\prime)$.
    
    To approximate such a hard constraint with a back-propagation-based method, we propose a novel loss function named \emph{Marginal Adaptation Loss} to constrain the update of the parameters of the encoder. The Marginal Adaptation Loss can be formulated as:
    \begin{equation}
    	\begin{aligned}
    		{{\cal L}_{m}} &= \frac{1}{\|\mathcal{D}_T\|}\sum_{i = 1}^{\|\mathcal{D}_T\|} {{{ \max (\|{f^\prime(\bm{x}_i;\theta_e^\prime)}} - {f(\bm{x}_i;\theta_e)}\|_F^2  - \sigma, 0)} }.
    	\end{aligned}
    	\label{eq:margin}
    \end{equation}

    The update of parameters of the classifier $q(\cdot)$ and encoder is guided by the entropy on the target domain. Based on the extracted representations $f^\prime(\bm{x}; \theta_e^\prime)$, we use a linear layer to work as a classifier and obtain the classification probability $\bm{p} = \texttt{softmax}(q^\prime(f^\prime(\bm{x}, \theta_e^\prime)))$ using a \texttt{softmax} operation. Then we take the entropy as the loss function to derive the gradients for updating the classifier and encoder, through which we can introduce the probabilistic distribution of target domains to our classifier: 
        \begin{equation}
        {{\cal L}_{e}} = - \frac{1}{N_b}\sum\nolimits_{i = 1}^{N_b}\sum\nolimits_{c=1}^C  \bm{p}_c \log \bm{p}_c .
        \label{eq:cls}
    \end{equation}

    \subsection{Differentiable Memory Bank}~\label{sec:method:dyn}
With Marginal Generalization, we are able to learn a well-adapted encoder that can extract discriminative features on the target domain. However, since there is no labeled data on the target domain, only training with the unsupervised losses $\mathcal{L}_m$ and $\mathcal{L}_e$ is hard to get a classifier $q(\cdot)$ with high performance on the target. To mitigate this issue, 
we propose to update the classifier with a differentiable memory bank. We utilize the memory bank to select prototypes suitable for the new domain, develop class-wise prototypes directly differentiable with loss function, and update the whole bank in every forward step.

    \textbf{Class-wise prototypes} are stored in the memory bank in order to enhance the classifier. Specifically, for each class $j$, the prototype $\bm{v}_j$ is initialized with the corresponding weights of the source classifier layer. In the self-supervised adaptation process, for each target sample $\bm{x}$, we extract the representations $f^\prime(\bm{x})$ and obtain the output of classifier $q^\prime(f^\prime(\bm{x}; \theta^\prime); \omega)$. Then we predict pseudo labels $\hat{y} = \arg \max [\texttt{softmax}(q^\prime(f^\prime(\bm{x})))]$ and utilize the entropy between representations and pseudo labels as the criterion to select the Top-$K$ instances of each class with highest classification confidence, where $K$ is a pre-defined hyper-parameter. After that, we utilize the representations of the Top-$K$ samples 
    to produce the class-wise prototypes $\bm{v}_j=\frac{1}{K} \sum_{i=1}^K f^\prime(\bm{x_i})$. 

    \textbf{Memory bank} is set to store the prototypes of each class $\mathcal{M}=\bigcup_{j=1}^C\{\bm{v}_j\}, \bm{v}_j\in \mathbb{R}^{d}$, where $\mathcal{M}$, $C$, and $d$ denote the memory bank, the number of classes, and feature dimension. At each forward step, we compute the prototypes, which will be further used to update the classifier weights $\omega$. For a given sample $\bm{x}$ with feature $\bm{z} = f^\prime(\bm{x};\theta)$, the classification probability of class $j$ can be computed as:
    \begin{equation}
        \bm{p}_j = \frac{\exp(\bm{z}\cdot\omega^j)}{\sum_k\exp(\bm{z}\cdot\omega^k)}, \,\,\,\, \omega^k \in \mathbb{R}^d
    \end{equation}
    where $\omega^j$ is the $j$-th element of the weight matrix $\omega$. Note that for $q(\cdot;\omega)$ to classify target samples correctly, the weight $\omega^j$ needs to be representative of features of the corresponding class $j$. This indicates that the meaning of $\omega^j$ coincides with the ideal cluster prototype of class $j$ in the target domain. Thus, we propose to use the estimate of the ideal target cluster prototypes $\{\bm{v}_j\}_{j=1}^C$ to update the classifier weights: $\omega^j\leftarrow \bm{v}_j$. This process is essential in learning a robust classifier for the target domain with no labeled data. 
    
\subsection{UniDG Learning for Domain Generalization}
In UniDG framework, the marginal generalization is proposed to learn a well-adapted feature encoder without catastrophic forgetting, and the differentiable memory bank is proposed to learn a discriminative classifier for the target domain. 
While updating $\omega$ with target prototypes, the overall learning objective is: 
\begin{equation}
    \mathcal{L}_\text{UniDG} = \mathcal{L}_e + \lambda \cdot \mathcal{L}_m 
\end{equation}


\section{Experiments} ~\label{sec:exp}
\subsection{Setup} ~\label{sec:exp:setup}
\textbf{\noindent{Dataset}} 
VLCS~\citep{torralba2011unbiased} contains 10,729 instances of 5 classes derived from four photographic datasets in accordance with different domains. PACS~\citep{li2017deeper} comprises four domains: art, cartoons, photos, and sketches, including 9,991 instances of classes. OfficeHome ~\citep{venkateswara2017deep} derives from domains like art, clipart, product, and real, containing 15,588 images of 65 classes. TerraIncognita~\citep{beery2018recognition}is a real-world dataset that collects photos of wild animals taken by cameras at different locations. It contains 24,788 photos of 10 classes according to the species of animals. DomainNet~\citep{peng2019moment} is the largest dataset for domain generalization tasks, including 6 domains, 345 classes, and a total of 586,575 images. \par

\textbf{\noindent{Evaluation Metric}} We evaluate \Model by taking 3 parallel trials with random seeds to calculate means and standard errors of classification accuracy (\%) on 5 datasets. There are 22 different novel environments to evaluate the abilities of the network for generalization. We report detailed results for each environment in Appendix~\ref{sec:supp:exp}. \par

\textbf{\noindent{Implementation Details}} 
All experimental results are conducted on NVIDIA A100 GPUs. If not specified, we utilize ResNet-50~\citep{he2016deep} for extracting visual features and a single classifier for classification. On test-time benchmarks, we utilize ERM~\citep{vapnik1991principles} algorithm as our default method for training source models. We also follow default hyper-parameters of DomainBed~\citep{gulrajani2020search} like initial learning rate of $5\times 10^{-5}$, weight decay of $0.0$, batch size of $32$, holdout fraction of $0.2$, and $\sigma$ of 0.15 (see Appendix \S~\ref{sec:imple} for more discussions).
\subsection{Main Results} ~\label{sec:exp:main_results}
We report experimental results on the domain generalization (\S~\ref{sec:exp:main_results:dg}) and test-time adaptation benchmarks (\S~\ref{sec:exp:main_results:tta}). \Model delivers new state-of-the-art performances on such benchmarks.
\subsubsection{Domain Generalization Benchmarks} \label{sec:exp:main_results:dg}
\Model prominently achieves a brilliant performance on Domain generalization tasks. Table~\ref{tab:dg} shows the performances of the existing advanced approaches for DG tasks using different pre-training methods. The upper part of the table demonstrates that with ImageNet pre-training, \Model significantly outperforms various classic models and shows satisfactory stability. Specifically, it achieved an average accuracy of 68.5 on VLCS, PACS, OfficeHome, Terrain, and DomainNet, exceeding AdaNPC by +2.0\%, and the best results on VLCS, PACS, terrain, and DomainNet. The remaining part of Table~\ref{tab:dg} shows more results with large-scale CLIP and SWAG pre-training. Expectedly, the CLIP- and SWAG-trained models outperform the traditional ImageNet-trained ones. However, impressively, with only ImageNet pre-training, \Model outperforms the CAR-FT model with CLIP pre-training by 1.1\% in the average accuracy (79.6\% vs. 78.5\%). On the terrain data set with complex domain shift, the accuracy of \Model reached 62.4\%, outperforming CAR-FT by 0.5\%.
\begin{table*}[t]
	\centering
	\caption{Overall out-of-domain accuracies with train-validation selection criterion on the \textbf{DomainBed} benchmark. The best result
		is highlighted in \textbf{bold}. \Model achieves the best performances on PACS, VLCS, OfficeHome, TerraIncognita, and DomainNet datasets. }
  \vspace{-3mm}
	\label{tab:dg}
	\resizebox{1.0\linewidth}{!}{
		\begin{tabular}{l|c|c|cccccc} 
			\toprule
			Algorithm & Venue & Pretraining & PACS & VLCS & OfficeHome & TerraInc & DomainNet & Avg.\\
			\hline ERM~(ResNet-50)~\citep{vapnik1991principles} & & \multirow{15}{*}{ImageNet} & $85.7 \pm 0.5$ & $77.4 \pm 0.3$ & $67.5 \pm 0.5$ & $47.2 \pm 0.4$ & $41.2 \pm 0.2$ & $63.8$ \\ 
			DANN~\citep{ganin2016domain} & JMLR'16 & & $84.6 \pm 1.1$ & $78.7 \pm 0.3$ & $68.6 \pm 0.4$ & $46.4 \pm 0.8$ & $41.8 \pm 0.2$ & $64.0$ \\
			
			MMD~\citep{li2018domain} & CVPR'18 & & $85.0 \pm 0.2$ & $76.7 \pm 0.9$ & $67.7 \pm 0.1$ & $42.2 \pm 1.4$ & $39.4 \pm 0.8$  & $62.2$ \\
			IRM ~\citep{arjovsky2019invariant} & ArXiv'20 &  & $83.5 \pm 0.8$ & $78.5 \pm 0.5$ & $64.3 \pm 2.2$ & $47.6 \pm 0.8$ & $33.9 \pm 2.8$  & $61.6$ \\
			FISH~\citep{shi2021gradient} & ICLR'22 &  & $85.5 \pm 0.3$ & $77.8 \pm 0.3$ & $68.6 \pm 0.4$ & $45.1 \pm 1.3$ & $42.7 \pm 0.2$ & $63.9$ \\
			SWAD~\citep{cha2021swad} & NeurIPS'21 &  & $88.1 \pm 0.1$ & $79.1 \pm 0.1$ & $70.6 \pm 0.2$ & $50.0 \pm 0.3$ & $46.5 \pm 0.1$  &  $66.9$ \\
			ERM (ViT-S/16)~\citep{dosovitskiy2020image} & ICLR'21 &  & $86.2 \pm 0.1$ & $79.7 \pm 0.0$ & $72.2 \pm 0.4$ & $42.0 \pm 0.8$ & $47.3 \pm 0.2$  & $65.5$ \\
			Fishr~\citep{rame2022fishr} & ICML'22&  & $85.5 \pm 0.2$ & $77.8 \pm 0.2$ & $68.6 \pm 0.2$ & $47.4 \pm 1.6$ & $41.7 \pm 0.0$ & $64.2$ \\
			MIRO\citep{cha2022domain} & ECCV'22 &  & $85.4 \pm 0.4$ & $79.0 \pm 0.0$ & $70.5 \pm 0.4$ & ${5 0 . 4} \pm {1 . 1}$ & $44.3 \pm 0.2$ & $65.9$ \\
			GMoE-S/16~\citep{li2022sparse} & ICLR'23 &  & ${8 8 . 1} \pm {0 . 1}$ & ${8 0 . 2} \pm {0 . 2}$ & $\mathbf{7 4 . 2} \pm \mathbf{0 . 4}$ & ${48.5} \pm 0.4$ & ${4 8 . 7} \pm {0 . 2}$ & $67.9$ \\ 
                ITTA~\citep{chen2023improved} & CVPR'23 & & $83.8 \pm 0.3$  & $76.9 \pm 0.6$ & $62.0 \pm 0.2$ & $43.2 \pm 0.5$ & $34.9 \pm 0.1$ & 60.2 \\
                DomainAdaptor~\citep{zhang2023domainadaptor} & ICCV'23 & & $84.9 \pm 0.2$ & $78.50 \pm 0.2$ & $66.7 \pm 0.3$ & - & - & - \\ 
                AdaNPC~\citep{zhang2023adanpc} & ICML'23 &  & $88.9 \pm 0.1$ & $80.2 \pm 0.2$ & $66.3 \pm 0.1$ & $\mathbf{54.0 \pm 0.1}$  & $43.1 \pm 0.8$  & 66.5 \\
			\rowcolor{mygray}
                \Model & Ours & & $\mathbf{89.0 \pm 0.3}$  
			& $\mathbf{81.6 \pm 0.1} $               & $ {68.9 \pm 0.1}$              & $ {52.9 \pm 0.2}$         & $ \mathbf{50.2 \pm 0.1}$  & $\mathbf{68.5}$ \\
			\midrule 
                \multicolumn{9}{c}{ \textit{UniDG with Existing DG Methods} } \\ \hline
                CORAL~\citep{sun2016deep} & ECCV'16 & {ImageNet} & $86.0 \pm 0.2$ & $77.7 \pm 0.5$ & $68.6 \pm 0.4$ & $46.4 \pm 0.8$ & $41.8 \pm 0.2$  & $64.1$  \\ 
                \rowcolor{mygray}
                \Model + CORAL &Ours & {ImageNet} & ${89.2 \pm 0.2}$  
			& ${82.1 \pm 0.1} $               & $ {70.6 \pm 0.1}$              & $ {53.0 \pm 0.2}$         & $ {51.3 \pm 0.1}$  & ${69.3}$ \textcolor{mygreen}{+5.2} \\ \hline 
                MIRO\citep{cha2022domain} & ECCV'22 &  {ImageNet} & $85.4 \pm 0.4$ & $79.0 \pm 0.0$ & $70.5 \pm 0.4$ & ${5 0 . 4} \pm {1 . 1}$ & $44.3 \pm 0.2$ & $65.9$ \\ 
                \rowcolor{mygray}
                \Model + MIRO & Ours &  {ImageNet} & $90.4 \pm 0.4$ & $84.1 \pm 0.2$ & $72.5 \pm 0.4$ & ${54.4} \pm {0.5}$ & $52.6 \pm 0.2$ & $70.8$ \textcolor{mygreen}{+4.9} \\ 
                \midrule
			\multicolumn{9}{c}{ViT-B/16~\citep{dosovitskiy2020image} Backbone} \\ \hline
			ERM~\citep{vapnik1991principles}  & & & 93.7 & 82.7 & 78.5 & 52.3 & 53.8 & 72.2 \\
			MIRO~\citep{cha2022domain} & ECCV'22 & \multirow{3}{*}{CLIP}& 95.6 & 82.2 & 82.5 & 54.3 & 54.0 & 73.7 \\
			DPL~\citep{zhang2021domain} & Arxiv'22 &   & 97.3 & 84.3 & 84.2 & 52.6 & 56.7 & 75.0 \\ 
			CAR-FT~\citep{mao2022context} & Arxiv'22 &   & 96.8 & 85.5 & 85.7 & 61.9 & \textbf{62.5} & 78.5 \\ \hline
                \rowcolor{mygray}
                UniDG & Ours &  & $96.7 \pm 0.4$ & $86.3 \pm 0.2$ & $86.2 \pm 0.1$ & $62.4 \pm 0.2 $ & $61.3 \pm 0.2$ & 78.6  \\ \hline
			\multicolumn{9}{c}{Base-scale Visual Backbone} \\ \hline
			
			ERM~\citep{vapnik1991principles}  & &  \multirow{2}{*}{SWAG} & $89.6 \pm 0.4$ & $78.6 \pm 0.3$ & $71.9 \pm 0.6$ & $51.4 \pm 1.8 $ & $48.5 \pm 0.6$ & 68.0 \\ 
			MIRO~\citep{cha2022domain}~(RegNetY-16GF~\citep{radosavovic2020designing}) & ECCV'22    &  & $\mathbf{97.4} \pm 0.2$ & $79.9 \pm 0.6$ & $80.4 \pm 0.2$ & $58.9 \pm 1.3 $ & $53.8 \pm 0.1$ & 74.1   \\ \hline
			\rowcolor{mygray}
			\Model + CORAL~\citep{sun2016deep} + ConvNeXt-B~\citep{liu2022convnet}  &Ours & ImageNet & $95.6 \pm 0.2$ & $\mathbf{84.5} \pm \mathbf{ 0.1}$ & $\mathbf{88.9} \pm \mathbf{0.3}$ & $\mathbf{69.6} \pm \mathbf{0.7} $ & $59.5 \pm 0.1$ & $\mathbf{ 79.6}$  \\ 
			\bottomrule
		\end{tabular}	
	}
	\vspace{-4mm}
\end{table*}
\begin{table*}[t]
	\begin{center}
		\vskip 0.10in
		\caption{Average accuracy ($\%$) using classifiers learned by ERM on the domain generalization benchmarks. We use ResNet-18/50 as backbones. \textbf{Bold} indicates the best for each benchmark.}
  \vspace{-3mm}
		\label{tab:tta}
		\resizebox{1.0\linewidth}{!}{
			\begin{tabular}{l|l|c|ccccc}
				\toprule
				Generalization Algorithm & Test-Time Algorithm  &Backbone& \text{VLCS} & \text{PACS} & \text{OfficeHome} & \text{TerraIncognita} & \text{Avg} \\ \midrule
				CLIP~\citep{radford2021learning} & Zero-Shot & ViT-B16 & 82.6$\pm$0.0 & 95.6$\pm$0.0 & 79.1$\pm$0.0 & 31.1$\pm$0.0 & 72.2\\ \hline
				\multirow{9}{*}{ERM~\citep{vapnik1991principles}}&  + None
				&\multirow{10}{*}{ResNet-18}& 74.9$\pm$0.5 & 79.3$\pm$0.8 & 62.1$\pm$0.3 & 40.6$\pm$1.2 & 64.2 \\
				& \text{+ PL}~\pub{ICMLW'13}~\citep{lee2013pseudo} && 63.0$\pm$2.7 & 71.0$\pm$1.8 & 58.2$\pm$3.2 & 37.4$\pm$7.2 & 57.4 \\
				& \text{+ PLClf}~\pub{ICMLW'13}~\citep{lee2013pseudo} && 74.9$\pm$0.6 & 78.1$\pm$2.3 & 61.9$\pm$0.4 & 41.8$\pm$1.9 & 64.2 \\
				& \text{+ SHOT}~\pub{ICML'20}~\citep{liang2020we} && 65.2$\pm$2.3 & 82.4$\pm$0.6 & 62.6$\pm$0.4 & 33.6$\pm$1.0 & 60.9 \\
				& \text{+ Tent}~\pub{ICLR'21}~\citep{wang2020tent} && 72.9$\pm$0.8 &  83.9$\pm$0.5 & 60.9$\pm$0.4 & 33.7$\pm$1.1 & 62.8 \\
				& \text{+ TentBN}~\pub{ICLR'21}~\citep{wang2020tent}  && 67.0$\pm$1.2 & 80.8$\pm$1.0 & 62.6$\pm$0.4 & 40.0$\pm$0.8 & 62.6 \\
				& \text{+ TentClf}~\pub{ICLR'21}~\citep{wang2020tent}  && 73.0$\pm$1.5 & 78.6$\pm$1.8 & 59.3$\pm$0.6 & 38.3$\pm$3.4 & 62.3 \\
				& \text{+ T3A}~\pub{NeurIPS'21}~\citep{iwasawa2021test} && 77.3$\pm$1.5 & 80.8$\pm$0.7 & 63.2$\pm$0.5 & 40.2$\pm$0.6 & 65.4 \\ 
				& \text{+ TAST}~\pub{ICLR'23}~\citep{jang2022test} && {77.3$\pm$0.7} & 81.9$\pm$0.4 & {63.7$\pm$0.5} & {42.6$\pm$0.7} & {66.4} \\ 
				\rowcolor{mygray}
				& \text{+ \Model}~\pub{Ours} &  & \textbf{80.9 $\pm$ 0.1}              & {81.7 $\pm$ 0.1}              & {58.4 $\pm$ 0.1}             & \textbf{47.9 $\pm$ 0.7}              &  \reshl{67.2}{0.8}                       \\
				\midrule
				\multirow{9}{*}{ERM~\citep{vapnik1991principles}}&  + None 
				&\multirow{10}{*}{ResNet-50}& 76.7$\pm$0.5 & 83.2$\pm$1.1 & 67.1$\pm$1.0 & 45.9$\pm$1.3 & 68.3 \\
				& \text{+ PL}~\pub{ICMLW'13}~\citep{lee2013pseudo} && 69.4$\pm$3.1 & 81.7$\pm$4.6 & 62.9$\pm$3.1 & 38.1$\pm$2.4 & 63.0 \\
				& \text{+ PLClf}~\pub{ICMLW'13}~\citep{lee2013pseudo} && 75.7$\pm$0.9 & 83.3$\pm$1.6 & 67.0$\pm$1.0 & 46.7$\pm$2.1 & 68.2 \\ 
				& \text{+ SHOT}~\pub{ICML'20}~\citep{liang2020we} & &67.1$\pm$0.9 & 84.1$\pm$1.2 & 67.7$\pm$0.7 & 35.2$\pm$0.8 & 63.5 \\ 
				& \text{+ Tent}~\pub{ICLR'21}~\citep{wang2020tent} && 73.0$\pm$1.3 & 85.2$\pm$0.6 & 66.3$\pm$0.8 & 37.1$\pm$2.0 & 65.4 \\ 
				& \text{+ TentBN}~\pub{ICLR'21}~\citep{wang2020tent} && 69.7$\pm$1.2 & 83.7$\pm$1.2 & 67.9$\pm$0.9 & 43.9$\pm$1.3 & 66.3 \\
				& \text{+ TentClf}~\pub{ICLR'21}~\citep{wang2020tent} && 75.8$\pm$0.7 & 82.7$\pm$1.6 & 66.8$\pm$1.0 & 43.6$\pm$2.6 & 67.2 \\ 
				& \text{+ T3A}~\pub{NeurIPS'21}~\citep{iwasawa2021test} && 77.3$\pm$0.4 & 83.9$\pm$1.1 & 68.3$\pm$0.8 & 45.6$\pm$1.1 & 68.8 \\
				& \text{+ TAST}~\pub{ICLR'23}~\citep{jang2022test} && {77.7$\pm$0.5} & 84.1$\pm$1.2 & 68.6$\pm$0.7 & {47.4$\pm$2.1}  & {69.5} \\
				\rowcolor{mygray}
				& \text{+ \Model}~\pub{Ours} & & \textbf{81.6 $\pm$ 0.1}   & \textbf{89.0 $\pm$ 0.3}              & \textbf{68.9 $\pm$ 0.1}              & \textbf{52.9 $\pm$ 0.2}              & \reshl{73.1}{3.6}                        \\
				\bottomrule 
			\end{tabular} \label{table:main_erm} 
		}
	\end{center}
	\vspace{-3mm}
\end{table*}  

\subsubsection{Test-Time Adaptation Benchmarks} \label{sec:exp:main_results:tta}
\Model remarkably outperforms all existing test-time methods including the state-of-the-art method, TAST~\citep{jang2022test}. Specifically, as shown in Table~\ref{tab:tta}, we choose ResNet-18 and ResNet-50 as the backbone and average accuracy as the metric to evaluate several test-time methods. \Model achieves an average accuracy of 67.2\% with ResNet-18 on VLCS, PACS, OfficeHome, and terrain, which is 0.8\% higher than the best-performing test-time method. The superiority of \Model is even more significant with ResNet-50: \Model achieves an average accuracy of 73.1\% on four benchmarks, largely exceeding the last state of the art, TAST~\citep{jang2022test}, by 3.5\%.

Except for ResNet-18 and ResNet-50, we further use UniDG with 12 mainstream backbones including CNN, MLP, and transformer architectures, and report the results in Figure~\ref{fig:backbone}. It turns out that \Model can significantly improve the performance of all the 12 backbones so that we conclude \Model is a universal architecture-agnostic method. Notably, the number of parameters of these models ranges from 1.59M to 303M, but \Model can significantly and consistently improve the performance by 5.4\% on average.
\begin{table*}[t]
	\centering
	\caption{Domain generalization accuracy with different backbone networks. \Model improves the
		performance agnostic to visual backbones. \textbf{Bold} type indicates performance improvement.}
  \vspace{-3mm}
	\label{tab:backbone}
	\resizebox{1.\linewidth}{!}{
		\begin{tabular}{l|c|c|cccc|l}
			\hline Type & Backbone & Method & VLCS & PACS & OfficeHome & Terra & Avg \\
			\hline \multirow{6}{*}{Light-weight Networks}
			& \multirow{2}{*}{ResNet-18~\citep{he2016deep}} & ERM & $ 76.5 \pm 0.1 $    & $ 79.2 \pm 0.1 $    & $ 56.0 \pm 0.1 $   & $ 40.3 \pm 0.0 $     & $ 63.0$                                     \\
			& & + \Model & \reshl{80.9 $\pm$ 0.1}{4.4}              & \reshl{81.7 $\pm$ 0.1}{2.5}              & \reshl{58.4 $\pm$ 0.1}{2.4}              & \reshl{47.9 $\pm$ 0.7}{7.6}              & \reshl{67.2}{4.2}\\ \cline{2-8}
			
			& \multirow{2}{*}{MobilenetV3~\citep{howard2019searching}} & ERM
			& $ 65.5\pm 0.2 $ & $ 79.1\pm 0.0 $ & $ 60.8\pm 0.2 $ & $ 30.4\pm 0.1 $ & $58.9$ \\    
			& & + \Model & \reshl{76.2 $\pm$ 0.1}{10.7}              & \reshl{85.3 $\pm$ 0.4}{6.2}              & \reshl{65.1 $\pm$ 0.2}{4.3}             & \reshl{34.7 $\pm$ 0.2}{4.3}             & \reshl{65.3}{6.4} \\ \cline{2-8}
			
			& \multirow{2}{*}{EfficientNetV2~\citep{tan2021efficientnetv2}} & ERM 
			& $ {69.9} \pm  {0.2}$ & $ {89.2} \pm  {0.0}$ & $ {73.6} \pm  {0.2}$ & $ {36.0} \pm  {0.2}$ & $67.2$ \\
			& & + \Model       & \reshl{78.6 $\pm$ 0.2}{8.7}              & \reshl{90.9 $\pm$ 0.1}{1.7}              & \reshl{77.2 $\pm$ 0.1}{3.6}              & \reshl{41.7 $\pm$ 0.4}{5.7}             & \reshl{72.1}{4.9}                       \\		
			\hline \multirow{6}{*}{Convolution Networks} & \multirow{2}{*}{ResNet-50~\citep{he2016deep}} & ERM
			& $ 77.1\pm 0.1 $ & $ 82.9\pm 0.1 $ & $ 65.2\pm 0.1 $ & $ 45.4\pm 0.1 $ & 67.6\\
			&  & + \Model & \reshl{81.6 $\pm$ 0.1}{4.5}              & \reshl{89.0 $\pm$ 0.3}{6.1}              & \reshl{68.9 $\pm$ 0.1}{3.7}              & \reshl{52.9 $\pm$ 0.2}{7.5}              & \reshl{73.1}{5.5}  \\
			\cline{2-8} & \multirow{2}{*}{ResNet-101~\citep{he2016deep}} & ERM 
			& $ 76.4 \pm 0.1 $ & $ 86.1\pm 0.0 $ & $ 67.4\pm 0.1 $ & $ 42.7\pm 0.1 $ & $68.1$ \\ 
			&  & + \Model & \reshl{80.5 $\pm$ 0.2}{4.1}              & \reshl{88.3 $\pm$ 0.1}{2.2}              & \reshl{70.3 $\pm$ 0.2}{2.9}              & \reshl{50.0 $\pm$ 0.5}{7.3}              & \reshl{72.3}{4.2} \\
			\cline{2-8} & \multirow{2}{*}{ConvNeXt-B~\citep{liu2022convnet}} & ERM 
			& $ 79.4\pm 0.0 $ & $ 92.7\pm 0.1 $ & $ 85.9\pm 0.1 $ & $ 60.9\pm 0.0 $ & $79.7$ \\ 
			&  & + \Model
			& \reshl{85.8 $\pm$ 0.3}{6.4}              & \reshl{95.3 $\pm$ 0.2}{2.6}              & \reshl{88.5 $\pm$ 0.0}{2.6}              & \reshl{65.3 $\pm$ 0.3}{4.4}              & \reshl{83.7}{4.0}                         \\
			
			
			\hline \multirow{10}{*}{Transformer Networks}  & \multirow{2}{*}{ViT-B16~\citep{dosovitskiy2020image}} & ERM 
			& $ 78.4\pm 0.1 $ & $ 80.3\pm 0.1 $ & $ 75.6\pm 0.1 $ & $ 43.6\pm 0.0 $ & $69.5$ \\ 
			&  & + \Model & \reshl{83.6 $\pm$ 0.1}{5.0}              & \reshl{85.4 $\pm$ 0.5}{5.1}              & \reshl{81.0 $\pm$ 0.0}{5.4}              & \reshl{51.4 $\pm$ 0.2}{8.0}              & \reshl{75.4}{5.9} \\
			
			\cline{2-8} & \multirow{2}{*}{ViT-L16~\citep{dosovitskiy2020image}} & ERM 
			& $ 76.4\pm 0.1 $ & $ 91.2\pm 0.1 $ & $ 83.3\pm 0.0 $ & $ 45.5\pm 0.0 $ & $74.1$ \\  
			&  & + \Model 
			& \reshl{83.2 $\pm$ 0.2}{6.8}              & \reshl{95.2 $\pm$ 0.1}{4.0}              & \reshl{87.5 $\pm$ 0.2}{4.2}              & \reshl{53.9 $\pm$ 0.4}{8.4}              & \reshl{79.9}{5.8}                        \\
			
			\cline{2-8} & \multirow{2}{*}{Hybrid ViT~\citep{dosovitskiy2020image}} & ERM 
			& $ 79.1\pm 0.1 $ & $ 89.1\pm 0.1 $ & $ 79.6\pm 0.1 $ & $ 52.9\pm 0.1 $ & $75.4$ \\  
			&  & + \Model 
			& \reshl{83.5 $\pm$ 0.1}{4.4}              & \reshl{93.5 $\pm$ 0.1}{4.4}              & \reshl{81.3 $\pm$ 0.1}{1.7}              & \reshl{60.1 $\pm$ 0.4}{7.2}              & \reshl{79.6}{4.2}      
			\\
			\cline{2-8} & \multirow{2}{*}{DeiT~\citep{touvron2021training}} & ERM 
			& $ 79.5\pm 0.1 $ & $ 88.0\pm 0.1 $ & $ 77.0\pm 0.1 $ & $ 49.5\pm 0.1 $ & $73.5$ \\ 
			&  & + \Model  
			& \reshl{85.1 $\pm$ 0.1}{5.6}              & \reshl{92.6 $\pm$ 0.3}{4.7}              & \reshl{79.5 $\pm$ 0.1}{2.5}              & \reshl{54.1 $\pm$ 0.4}{4.8}              & \reshl{77.8}{4.4} \\
			
			\cline{2-8} & \multirow{2}{*}{Swin Transformer~\citep{liu2021swin}} & ERM 
			& $ 80.0\pm 0.1 $ & $ 90.2\pm 0.1 $ & $ 81.6\pm 0.1 $ & $ 57.0\pm 0.0 $ & $77.2$ \\ 
			&  & + \Model 
			& \reshl{85.0 $\pm$ 0.1}{5.0}              & \reshl{94.3 $\pm$ 0.2}{4.1}              & \reshl{84.6 $\pm$ 0.1}{3.0}              & \reshl{62.0 $\pm$ 0.3}{5.0}              & \reshl{81.5}{4.3}                       \\
			\hline 
			\multirow{4}{*}{Multi-Layer Perceptron}& \multirow{2}{*}{Mixer-B16~\citep{tolstikhin2021mlp}} & ERM 
			& $ 73.6\pm 0.1 $ & $ 75.8\pm 0.0 $ & $ 52.4\pm 0.1 $ & $ 26.8\pm 0.1 $ & $57.2$ \\ 
			&  & +\Model 	   & \reshl{81.3 $\pm$ 0.2}{7.7}              & \reshl{82.3 $\pm$ 0.1}{6.5}              & \reshl{57.7 $\pm$ 0.3}{5.2}              & \reshl{41.2 $\pm$ 0.5}{14.4}             & \reshl{65.6}{8.4}                        \\
			\cline{2-8} & \multirow{2}{*}{Mixer-L16~\citep{tolstikhin2021mlp}} & ERM 
			& $ 77.1\pm 0.1 $ & $ 85.0\pm 0.1 $ & $ 70.3\pm 0.1 $ & $ 36.6\pm 0.0 $ & $67.4$ \\ 
			&  & + \Model 	   & \reshl{83.0 $\pm$ 0.1}{4.9}              & \reshl{88.5 $\pm$ 0.2}{3.5}              & \reshl{75.6 $\pm$ 0.1}{5.3}              & \reshl{45.0 $\pm$ 1.4}{8.4}              & \reshl{73.0}{5.6} \\			
			\hline
		\end{tabular}
	}
	\vspace{-2mm}
\end{table*}

\subsection{Ablation Study} ~\label{sec:exp:ablation_study}

\textbf{Effectiveness of Marginal Generalization.}
Table~\ref{tab:ablation} shows Marginal Generalization significantly improves the performance on target domains compared with the baseline model by +3.3\% (70.9\% vs. 67.6\%). With the classifier adaptation scheme (\S~\ref{sec:method:dyn}) but no Marginal Generalization, the performance reaches 70.8\%, bringing a +3.2\% improvement. While further integrating the two schemes, the ability of the network for domain generalization gets significantly boosted, increasing from 67.6\% to 71.9\%.

\textbf{ Effectiveness of Differentiable Memory Bank}
As shown in the 4th, 5th, and 7th rows of Table~\ref{tab:ablation}, Differentiable Memory Bank~(\S~\ref{sec:method:dyn}) also significantly improves the generalization ability of the model. Referring to the 4th row, the memory bank effectively boosts the performance of the base model from 67.6\% to 70.4\% (+2.8\%). 
\begin{wrapfigure}{r}{7.3cm}
	\begin{minipage}{0.52\textwidth}
		\begin{center}
        \vspace{-7mm}
        \begin{table}[H]
            \centering
            \caption{\textbf{Ablation Study}. We take the mean accuracy (mAcc) on the PACS, VLCS, OfficeHome, and TerraInc datasets as the evaluation metric.}
        \vspace{-3mm}
        \label{tab:ablation}
        \resizebox{1.0\linewidth}{!}{
        \begin{tabular}{cccc|c|c}
            \hline \multicolumn{4}{c}{Ablation Components} & mAcc  & $\Delta$ \\
            \hline
            $\mathcal{L}_{m}$~~(Eq.~\ref{eq:margin}) & $\mathcal{L}_{e}$~~(Eq
            .~\ref{eq:cls}) 
            & $\mathcal{M}$ & $\omega\leftarrow v$ & (\%) & (\%)\\
            \hline 
            \ding{56} & \ding{56} & \ding{56} & \ding{56} 
            & $67.6 \pm 0.1 $ & $-$ \\ 
            \hline
            \ding{51} & \ding{56} & \ding{56} & \ding{56}
            & $ 70.9 \pm 0.2$ & $\mathbf{+ 3.3}$ \\
            \ding{56} & \ding{51} & \ding{56} & \ding{56}
            & $70.8 \pm 0.1$ & $\mathbf{+ 3.2}$ \\
            \ding{56} & \ding{56} & \ding{51} & \ding{56}
            & $ 70.4 \pm 0.2$ & $\mathbf{+ 2.8}$ \\
            \ding{56} & \ding{56} & \ding{56} & \ding{51}
            & $ 70.7 \pm 0.1 $ & $\mathbf{+ 3.1}$ \\
            \hline
            \ding{51} & \ding{51} & \ding{56} & \ding{56}
            & $71.9 \pm 0.3$ & $\mathbf{+ 4.3}$ \\
            \ding{56} & \ding{56} & \ding{51} & \ding{51}
            & $71.6 \pm 0.0$ & $\mathbf{+ 4.0}$ \\
            \hline
            \ding{51} & \ding{51} & \ding{51} & \ding{51}
            & $73.1 \pm 0.2 $ & $\mathbf{+ 5.5 }$  \\
            \hline
        \end{tabular}
        }
        \end{table}
		\end{center}
		\vspace{-15mm}
	\end{minipage}
\end{wrapfigure}
Meanwhile, when combining differentiable memory bank and Marginal Generalization, a further improvement of +5.5\% can be achieved. It reveals that the proposed schemes can be mutually beneficial, where the adapted model has refined gradients and a differentiable memory bank receives better prototypes. Thus they enhance the ability of networks to generalization together.

\subsection{Quantitative Analysis} ~\label{sec:exp:vis}
\begin{figure*}[t]
	\centering
	\caption{Accuracy accumulation curves on VLCS. \Model outperforms the base ERM model by about 5\% in accuracy. Note we randomly select \textbf{10 different trial seeds} for better comparison.}
	\vspace{-3mm}
	\label{fig:curve}
	{\includegraphics[width=0.24\linewidth]{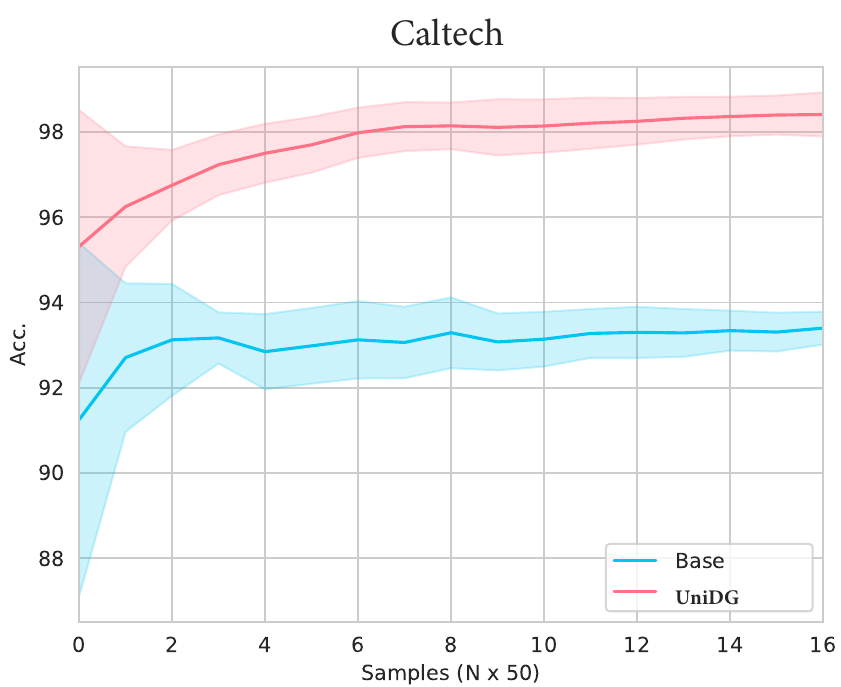}}
	{\includegraphics[width=0.24\linewidth]{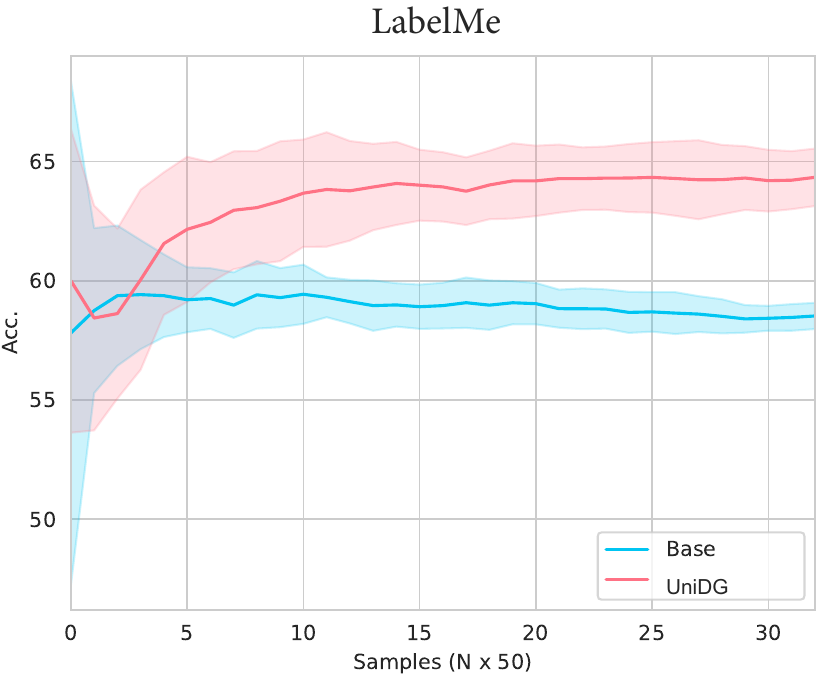}}
	{\includegraphics[width=0.24\linewidth]{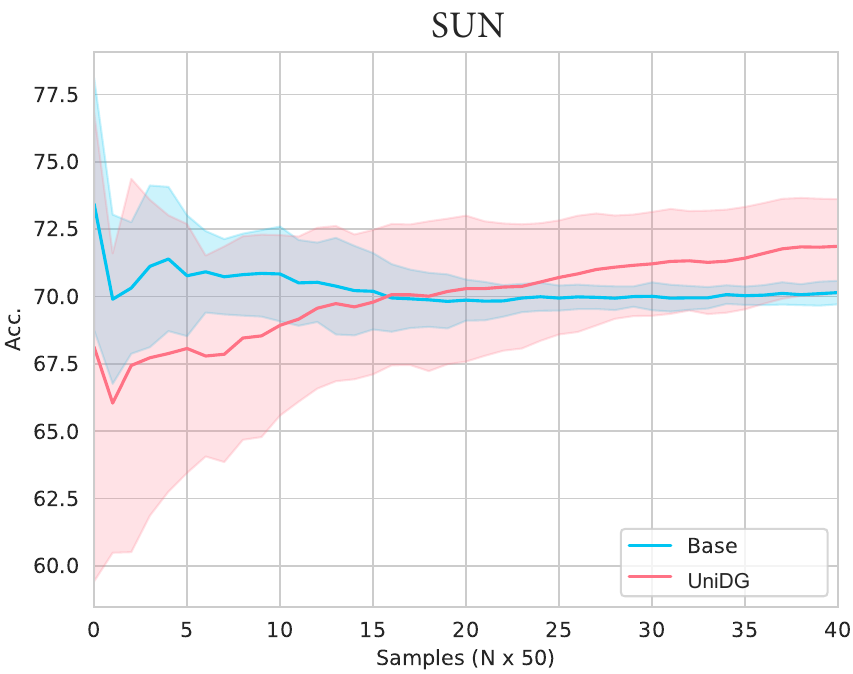}}
	{\includegraphics[width=0.24\linewidth]{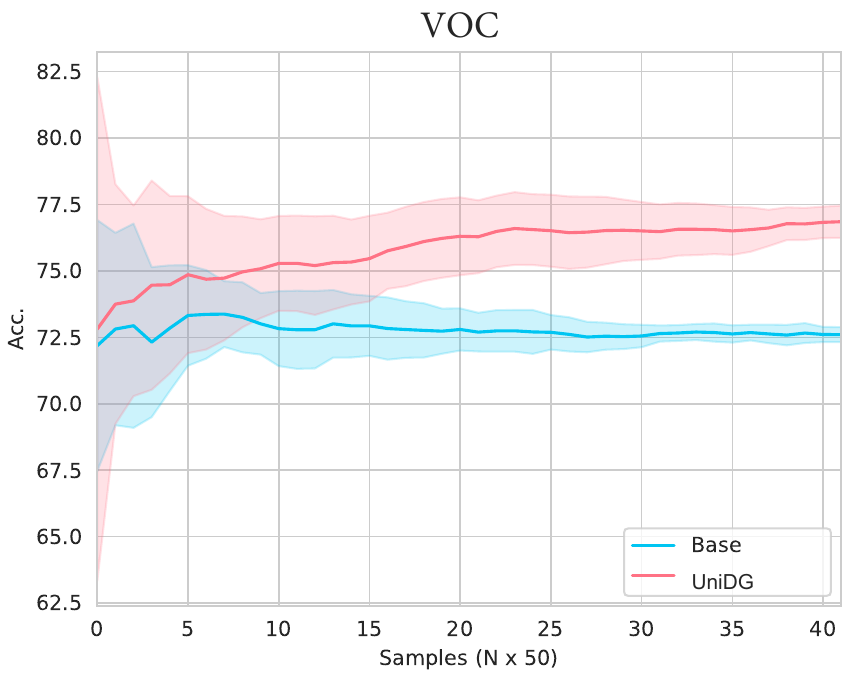}}
 \vspace{-3mm}
\end{figure*}
Figure~\ref{fig:curve} shows the accumulation curves of each instance interval across four domains on the VLCS~\citep{li2017deeper} dataset by 10 parallel trials. \Model brings significant and stable improvements on each domain, for which the fluctuation range of accumulation accuracy is close to the base model and mean scores are prominently improved.

\begin{table}[ht]
    \vspace{-3mm}
    \caption{Source knowledge preserve and training efficiency of UniDG.}
    \label{tab:source_preserve}
    \centering
    \vspace{-3mm}	
    \subfloat[Source Knowledge Preserve~\label{tab:source}]{%
    \resizebox{0.45\linewidth}{!}{
    \begin{tabular}{lccc}
    \toprule
    VLCS & L & S & V \\ \midrule
    Source model       & 96.02 & 97.14 & 98.33 \\
    TENT  & 92.15 \textcolor{mygreen}{$-$3.9}  & 94.23 \textcolor{mygreen}{$-$2.9} & 95.13 \textcolor{mygreen}{$-$3.2} \\
    UniDG        & 94.32 \textcolor{mygreen}{$-$1.7} & 96.68 \textcolor{mygreen}{$-$0.4} & 97.63 \textcolor{mygreen}{$-$0.7} \\
    \bottomrule
\end{tabular}		}
    }
    \subfloat[Efficiency of UniDG~\label{tab:efficiency}]{
    \resizebox{0.40\linewidth}{!}{
    \begin{tabular}{lcc}
    \toprule
    Method            & Wall Clock Time (s)       \\ \midrule
    TENT (Res50)  &   0.581           \\
    UniDG (Res50) &   0.587  \\
    \bottomrule
    \end{tabular}}
} 
\vspace{-3mm}	
\end{table}
 As shown in Table~\ref{tab:source_preserve}, 1) referring to Table~\ref{tab:source}, we observe a smaller performance decrease of UniDG after adaption on the source domains. It proves that UniDG can better preserve pretrained source knowledge. 2) In Table~\ref{tab:efficiency}, we detail the training efficiency of UniDG and compare our method with TENT on wall clock time with the NVIDIA A100 GPU. It reveals that although we propose to update the parameters of the whole network, the computation burden will not sharply increase.
\subsection{Commonality Analysis} ~\label{sec:exp:backbone}
1) \textbf{Light-weight Networks}
\Model brings out significant average improvements of 5.1\% on light-weight MobileNet V3~\citep{howard2019searching}, EfficientNet V2~\citep{tan2021efficientnetv2}, and ResNet 18~\citep{he2016deep}. For example, the accuracy of MobileNet V3 has been improved by as much as 6.4\%, which proves the strong feasibility of \Model to improve the performance of edge devices for generalizing in unseen environments.
2) \textbf{Architecture-Free}
\Model is a unified solution based on online adaptation to handle domain shifts. As shown in Table~\ref{tab:backbone}, \Model has a general improvement of about 5\% on 10+ mainstream visual networks including CNN, Transformer, and MLP as their backbones. The highest improvement comes from Mixer-B16~\citep{tolstikhin2021mlp}, which increased from 57.2\% to 65.6\%.
\vspace{-3mm}

\section{Related Work} ~\label{sec:related}
\textbf{Domain Generalization} Domain Generalization (DG) can be classified into three types:
1) \textbf{Representation Learning}: These methods extract specific features from source domains and assume them robust in target domains. One approach is domain alignment~\citep{li2018deep,li2018domain,dg_mmld}, extracting domain-invariant representations from source domains, which is a non-trivial task. Therefore feature disentanglement~\citep{rojas2018invariant,piratla2020efficient,christiansen2021causal,mahajan2021domain,sun2021recovering,liu2021learning}, loosens the constraint, learning disentangled representations.
2) \textbf{Foundation Models}: different backbones reveal the diverse ability to tackle the DG problem. These methods~\citep{li2017deeper,ding2017deep,carlucci2019domain,li2022sparse}  optimize the architecture of the mainstream backbone for DG. GMoE~\citep{li2022sparse} based on ViT~\citep{dosovitskiy2020image}, replaces the FFN layers with mixture-of-experts, allowing different experts to focus on the different visual attributes.
3) \textbf{Learning Strategy}: These methods utilize machine learning strategy to enhance the model’s generalization capability on target domains, including meta-learning and ensemble learning. Meta-learning~\citep{li2018learning,li2019feature,li2019episodic,dou2019domain,liu2020shape,chen2022ost,li2021metasaug} divide training data into meta-train and meta-test sets, then simulate domain shift and update parameters during training. Ensemble learning~\citep{ding2017deep,zhou2021domain,cha2021swad} learns model copies to extract features and migrate their ensemble to target domains.

\textbf{Continual Learning} Continual learning~\citep{de2021continual} aims to relieve continuous domain shifts, which face complicated catastrophic forgetting. Existing methods~\citep{rebuffi2017icarl,zenke2017continual,kirkpatrick2017overcoming,li2017learning,lao2020continuous} propose regularization and replay to reinforce learning representations space from parameters and data stream perspectives.
Recently,self-supervised learning~\citep{radford2015unsupervised,he2022masked,grill2020bootstrap} utilize prior knowledge obtained by pre-training with massive datasets and have shown strong performance in DG.~\cite{radford2021learning} trains image encoder and text encoder jointly, matching 400 million (image, text)pairs. Besides, researchers have noted the superiority of causal learning~\citep{zhou2021domain,mahajan2021domain} in domain generalization.

\textbf{Test-time Adaptation} TTA schemes~\citep{karani2021test,iwasawa2021test,sun2020test,park2023test} propose to update model parameters based on target data. 

1) \textbf{Adversarial Learning}: With the advancement of generative adversarial networks,~\cite{li2020model,yeh2021sofa,kurmi2021domain} generate target data with generative models, improving the ability to handle domain shift without the support of source data.2) \textbf{normalization-based}: The normalization method replaces the batch normalization (BN) statistics of the trained model with the BN statistics estimated on test data and updates parameters of the BN layers only, with the backbone network frozen.~\cite{wang2020tent} aims to minimize entropy during testing.~\cite{schneider2020improving} uses Wasserstein distance between source and target statistics as the measurement.
3) \textbf{Bayesian Learning}: \cite{zhou2021training} analyses TTA from a Bayesian perspective~\citep{li2016revisiting,hu2021mixnorm,you2021test} and proposes a regularized entropy minimization procedure achieved by approximating the probability density during the training time.


\section{Discussion and Conclusion } ~\label{sec:conclusion}
Aiming at the OOD problem, this paper proposes a general self-supervised online learning scheme, named UniDG, to update all the parameters of the model during the testing phase. Specifically, \Model contains Marginal Generalization and Differentiable Memory Bank, which can successfully balance the conservation of source knowledge and generalization ability to novel environments. Our method shows high effectiveness and potential for complex domain shifts in actual scenarios. On four domain generalization benchmarks, \Model achieved a new state-of-the-art performance with an average accuracy of 79.6\%. Additionally, \Model improved 12 backbone models by an average of 5.4\%. By comparing with existing pre-trained model and other test-time methods, we show it is a promising direction to develop the online adaptation method to deal with the OOD problem.

\bibliography{iclr2024_conference}

\begin{thebibliography}{92}
\providecommand{\natexlab}[1]{#1}
\providecommand{\url}[1]{\texttt{#1}}
\expandafter\ifx\csname urlstyle\endcsname\relax
  \providecommand{\doi}[1]{doi: #1}\else
  \providecommand{\doi}{doi: \begingroup \urlstyle{rm}\Url}\fi

\bibitem[Arjovsky et~al.(2019)Arjovsky, Bottou, Gulrajani, and Lopez-Paz]{arjovsky2019invariant}
Martin Arjovsky, L{\'e}on Bottou, Ishaan Gulrajani, and David Lopez-Paz.
\newblock Invariant risk minimization.
\newblock \emph{arXiv preprint arXiv:1907.02893}, 2019.

\bibitem[Beery et~al.(2018)Beery, Van~Horn, and Perona]{beery2018recognition}
Sara Beery, Grant Van~Horn, and Pietro Perona.
\newblock Recognition in terra incognita.
\newblock In \emph{Proceedings of the European conference on computer vision (ECCV)}, pp.\  456--473, 2018.

\bibitem[Carlucci et~al.(2019)Carlucci, D'Innocente, Bucci, Caputo, and Tommasi]{carlucci2019domain}
Fabio~M Carlucci, Antonio D'Innocente, Silvia Bucci, Barbara Caputo, and Tatiana Tommasi.
\newblock Domain generalization by solving jigsaw puzzles.
\newblock In \emph{CVPR}, pp.\  2229--2238, 2019.

\bibitem[Cha et~al.(2021)Cha, Chun, Lee, Cho, Park, Lee, and Park]{cha2021swad}
Junbum Cha, Sanghyuk Chun, Kyungjae Lee, Han-Cheol Cho, Seunghyun Park, Yunsung Lee, and Sungrae Park.
\newblock Swad: Domain generalization by seeking flat minima.
\newblock \emph{Advances in Neural Information Processing Systems}, 34:\penalty0 22405--22418, 2021.

\bibitem[Cha et~al.(2022)Cha, Lee, Park, and Chun]{cha2022domain}
Junbum Cha, Kyungjae Lee, Sungrae Park, and Sanghyuk Chun.
\newblock Domain generalization by mutual-information regularization with pre-trained models.
\newblock In \emph{Computer Vision--ECCV 2022: 17th European Conference, Tel Aviv, Israel, October 23--27, 2022, Proceedings, Part XXIII}, pp.\  440--457. Springer, 2022.

\bibitem[Chen et~al.(2022{\natexlab{a}})Chen, Li, Han, Liu, and Yu]{chen2022compound}
Chaoqi Chen, Jiongcheng Li, Xiaoguang Han, Xiaoqing Liu, and Yizhou Yu.
\newblock Compound domain generalization via meta-knowledge encoding.
\newblock In \emph{CVPR}, pp.\  7119--7129, 2022{\natexlab{a}}.

\bibitem[Chen et~al.(2022{\natexlab{b}})Chen, Wang, Darrell, and Ebrahimi]{chen2022contrastive}
Dian Chen, Dequan Wang, Trevor Darrell, and Sayna Ebrahimi.
\newblock Contrastive test-time adaptation.
\newblock In \emph{Proceedings of the IEEE/CVF Conference on Computer Vision and Pattern Recognition}, pp.\  295--305, 2022{\natexlab{b}}.

\bibitem[Chen et~al.(2022{\natexlab{c}})Chen, Zhang, Song, Liu, and Wang]{chen2022self}
Liang Chen, Yong Zhang, Yibing Song, Lingqiao Liu, and Jue Wang.
\newblock Self-supervised learning of adversarial example: Towards good generalizations for deepfake detection.
\newblock In \emph{CVPR}, pp.\  18710--18719, 2022{\natexlab{c}}.

\bibitem[Chen et~al.(2022{\natexlab{d}})Chen, Zhang, Song, Wang, and Liu]{chen2022ost}
Liang Chen, Yong Zhang, Yibing Song, Jue Wang, and Lingqiao Liu.
\newblock {OST}: Improving generalization of deepfake detection via one-shot test-time training.
\newblock In \emph{NeurIPS}, 2022{\natexlab{d}}.

\bibitem[Chen et~al.(2023)Chen, Zhang, Song, Shan, and Liu]{chen2023improved}
Liang Chen, Yong Zhang, Yibing Song, Ying Shan, and Lingqiao Liu.
\newblock Improved test-time adaptation for domain generalization.
\newblock In \emph{Proceedings of the IEEE/CVF Conference on Computer Vision and Pattern Recognition}, pp.\  24172--24182, 2023.

\bibitem[Christiansen et~al.(2021)Christiansen, Pfister, Jakobsen, Gnecco, and Peters]{christiansen2021causal}
Rune Christiansen, Niklas Pfister, Martin~Emil Jakobsen, Nicola Gnecco, and Jonas Peters.
\newblock A causal framework for distribution generalization.
\newblock \emph{IEEE TPAMI}, 2021.

\bibitem[De~Lange et~al.(2021)De~Lange, Aljundi, Masana, Parisot, Jia, Leonardis, Slabaugh, and Tuytelaars]{de2021continual}
Matthias De~Lange, Rahaf Aljundi, Marc Masana, Sarah Parisot, Xu~Jia, Ale{\v{s}} Leonardis, Gregory Slabaugh, and Tinne Tuytelaars.
\newblock A continual learning survey: Defying forgetting in classification tasks.
\newblock \emph{IEEE transactions on pattern analysis and machine intelligence}, 44\penalty0 (7):\penalty0 3366--3385, 2021.

\bibitem[Ding \& Fu(2017)Ding and Fu]{ding2017deep}
Zhengming Ding and Yun Fu.
\newblock Deep domain generalization with structured low-rank constraint.
\newblock \emph{IEEE Transactions on Image Processing}, 27\penalty0 (1):\penalty0 304--313, 2017.

\bibitem[Dosovitskiy et~al.(2020)Dosovitskiy, Beyer, Kolesnikov, Weissenborn, Zhai, Unterthiner, Dehghani, Minderer, Heigold, Gelly, et~al.]{dosovitskiy2020image}
Alexey Dosovitskiy, Lucas Beyer, Alexander Kolesnikov, Dirk Weissenborn, Xiaohua Zhai, Thomas Unterthiner, Mostafa Dehghani, Matthias Minderer, Georg Heigold, Sylvain Gelly, et~al.
\newblock An image is worth 16x16 words: Transformers for image recognition at scale.
\newblock \emph{arXiv preprint arXiv:2010.11929}, 2020.

\bibitem[Dou et~al.(2019)Dou, de~Castro, Kamnitsas, and Glocker]{dou2019domain}
Qi~Dou, Daniel~Coelho de~Castro, Konstantinos Kamnitsas, and Ben Glocker.
\newblock Domain generalization via model-agnostic learning of semantic features.
\newblock In \emph{NeurIPS}, pp.\  6447--6458, 2019.

\bibitem[Ebrahimi et~al.(2020)Ebrahimi, Meier, Calandra, Darrell, and Rohrbach]{ebrahimi2020adversarial}
Sayna Ebrahimi, Franziska Meier, Roberto Calandra, Trevor Darrell, and Marcus Rohrbach.
\newblock Adversarial continual learning.
\newblock In \emph{ECCV}. Springer, 2020.

\bibitem[Ganin et~al.(2016)Ganin, Ustinova, Ajakan, Germain, Larochelle, Laviolette, Marchand, and Lempitsky]{ganin2016domain}
Yaroslav Ganin, Evgeniya Ustinova, Hana Ajakan, Pascal Germain, Hugo Larochelle, Fran{\c{c}}ois Laviolette, Mario Marchand, and Victor Lempitsky.
\newblock Domain-adversarial training of neural networks.
\newblock \emph{The Journal of Machine Learning Research}, 17\penalty0 (1):\penalty0 2096--2030, 2016.

\bibitem[Grill et~al.(2020)Grill, Strub, Altch{\'e}, Tallec, Richemond, Buchatskaya, Doersch, Avila~Pires, Guo, Gheshlaghi~Azar, et~al.]{grill2020bootstrap}
Jean-Bastien Grill, Florian Strub, Florent Altch{\'e}, Corentin Tallec, Pierre Richemond, Elena Buchatskaya, Carl Doersch, Bernardo Avila~Pires, Zhaohan Guo, Mohammad Gheshlaghi~Azar, et~al.
\newblock Bootstrap your own latent-a new approach to self-supervised learning.
\newblock \emph{Advances in neural information processing systems}, 33:\penalty0 21271--21284, 2020.

\bibitem[Gulrajani \& Lopez-Paz(2020)Gulrajani and Lopez-Paz]{gulrajani2020search}
Ishaan Gulrajani and David Lopez-Paz.
\newblock In search of lost domain generalization.
\newblock In \emph{International Conference on Learning Representations}, 2020.

\bibitem[He et~al.(2016)He, Zhang, Ren, and Sun]{he2016deep}
Kaiming He, Xiangyu Zhang, Shaoqing Ren, and Jian Sun.
\newblock Deep residual learning for image recognition.
\newblock In \emph{CVPR}, pp.\  770--778, 2016.

\bibitem[He et~al.(2022)He, Chen, Xie, Li, Doll{\'a}r, and Girshick]{he2022masked}
Kaiming He, Xinlei Chen, Saining Xie, Yanghao Li, Piotr Doll{\'a}r, and Ross Girshick.
\newblock Masked autoencoders are scalable vision learners.
\newblock In \emph{Proceedings of the IEEE/CVF Conference on Computer Vision and Pattern Recognition}, pp.\  16000--16009, 2022.

\bibitem[Howard et~al.(2019)Howard, Sandler, Chu, Chen, Chen, Tan, Wang, Zhu, Pang, Vasudevan, et~al.]{howard2019searching}
Andrew Howard, Mark Sandler, Grace Chu, Liang-Chieh Chen, Bo~Chen, Mingxing Tan, Weijun Wang, Yukun Zhu, Ruoming Pang, Vijay Vasudevan, et~al.
\newblock Searching for mobilenetv3.
\newblock In \emph{Proceedings of the IEEE/CVF international conference on computer vision}, pp.\  1314--1324, 2019.

\bibitem[Hu et~al.(2021)Hu, Uzunbas, Chen, Wang, Shah, Nevatia, and Lim]{hu2021mixnorm}
Xuefeng Hu, Gokhan Uzunbas, Sirius Chen, Rui Wang, Ashish Shah, Ram Nevatia, and Ser-Nam Lim.
\newblock Mixnorm: Test-time adaptation through online normalization estimation.
\newblock \emph{arXiv preprint arXiv:2110.11478}, 2021.

\bibitem[Ioffe \& Szegedy(2015)Ioffe and Szegedy]{ioffe2015batch}
Sergey Ioffe and Christian Szegedy.
\newblock Batch normalization: Accelerating deep network training by reducing internal covariate shift.
\newblock In \emph{ICLR}, pp.\  448--456. PMLR, 2015.

\bibitem[Iwasawa \& Matsuo(2021)Iwasawa and Matsuo]{iwasawa2021test}
Yusuke Iwasawa and Yutaka Matsuo.
\newblock Test-time classifier adjustment module for model-agnostic domain generalization.
\newblock In \emph{NeuIPS}, 2021.

\bibitem[Jacot et~al.(2018)Jacot, Gabriel, and Hongler]{jacot2018neural}
Arthur Jacot, Franck Gabriel, and Cl{\'e}ment Hongler.
\newblock Neural tangent kernel: Convergence and generalization in neural networks.
\newblock \emph{Advances in neural information processing systems}, 31, 2018.

\bibitem[Jang \& Chung(2023)Jang and Chung]{jang2022test}
Minguk Jang and Sae-Young Chung.
\newblock Test-time adaptation via self-training with nearest neighbor information.
\newblock In \emph{International Conference on Learning Representations}, 2023.

\bibitem[Karani et~al.(2021)Karani, Erdil, Chaitanya, and Konukoglu]{karani2021test}
Neerav Karani, Ertunc Erdil, Krishna Chaitanya, and Ender Konukoglu.
\newblock Test-time adaptable neural networks for robust medical image segmentation.
\newblock \emph{Medical Image Analysis}, 68:\penalty0 101907, 2021.

\bibitem[Kirkpatrick et~al.(2017)Kirkpatrick, Pascanu, Rabinowitz, Veness, Desjardins, Rusu, Milan, Quan, Ramalho, Grabska-Barwinska, et~al.]{kirkpatrick2017overcoming}
James Kirkpatrick, Razvan Pascanu, Neil Rabinowitz, Joel Veness, Guillaume Desjardins, Andrei~A Rusu, Kieran Milan, John Quan, Tiago Ramalho, Agnieszka Grabska-Barwinska, et~al.
\newblock Overcoming catastrophic forgetting in neural networks.
\newblock \emph{Proceedings of the national academy of sciences}, 114\penalty0 (13):\penalty0 3521--3526, 2017.

\bibitem[Kurmi et~al.(2021)Kurmi, Subramanian, and Namboodiri]{kurmi2021domain}
Vinod~K Kurmi, Venkatesh~K Subramanian, and Vinay~P Namboodiri.
\newblock Domain impression: A source data free domain adaptation method.
\newblock In \emph{WACV}, pp.\  615--625, 2021.

\bibitem[Lao et~al.(2020)Lao, Jiang, Havaei, and Bengio]{lao2020continuous}
Qicheng Lao, Xiang Jiang, Mohammad Havaei, and Yoshua Bengio.
\newblock Continuous domain adaptation with variational domain-agnostic feature replay.
\newblock \emph{arXiv preprint arXiv:2003.04382}, 2020.

\bibitem[Lee(2013)]{lee2013pseudo}
Dong-Hyun Lee.
\newblock Pseudo-label: The simple and efficient semi-supervised learning method for deep neural networks.
\newblock In \emph{Workshop on Challenges in Representation Learning, ICML}, volume~3, pp.\ ~2, 2013.

\bibitem[Li et~al.(2023)Li, Shen, Yang, Wang, Ren, Che, Zhang, and Liu]{li2022sparse}
Bo~Li, Yifei Shen, Jingkang Yang, Yezhen Wang, Jiawei Ren, Tong Che, Jun Zhang, and Ziwei Liu.
\newblock Sparse mixture-of-experts are domain generalizable learners.
\newblock In \emph{International Conference on Learning Representations}, 2023.

\bibitem[Li et~al.(2017)Li, Yang, Song, and Hospedales]{li2017deeper}
Da~Li, Yongxin Yang, Yi-Zhe Song, and Timothy~M Hospedales.
\newblock Deeper, broader and artier domain generalization.
\newblock In \emph{ICCV}, pp.\  5542--5550, 2017.

\bibitem[Li et~al.(2018{\natexlab{a}})Li, Yang, Song, and Hospedales]{li2018learning}
Da~Li, Yongxin Yang, Yi-Zhe Song, and Timothy~M Hospedales.
\newblock Learning to generalize: Meta-learning for domain generalization.
\newblock In \emph{AAAI}, 2018{\natexlab{a}}.

\bibitem[Li et~al.(2019{\natexlab{a}})Li, Zhang, Yang, Liu, Song, and Hospedales]{li2019episodic}
Da~Li, Jianshu Zhang, Yongxin Yang, Cong Liu, Yi-Zhe Song, and Timothy~M Hospedales.
\newblock Episodic training for domain generalization.
\newblock In \emph{ICCV}, pp.\  1446--1455, 2019{\natexlab{a}}.

\bibitem[Li et~al.(2018{\natexlab{b}})Li, Pan, Wang, and Kot]{li2018domain}
Haoliang Li, Sinno~Jialin Pan, Shiqi Wang, and Alex~C Kot.
\newblock Domain generalization with adversarial feature learning.
\newblock In \emph{Proceedings of the IEEE conference on computer vision and pattern recognition}, pp.\  5400--5409, 2018{\natexlab{b}}.

\bibitem[Li et~al.(2020)Li, Jiao, Cao, Wong, and Wu]{li2020model}
Rui Li, Qianfen Jiao, Wenming Cao, Hau-San Wong, and Si~Wu.
\newblock Model adaptation: Unsupervised domain adaptation without source data.
\newblock In \emph{CVPR}, pp.\  9641--9650, 2020.

\bibitem[Li et~al.(2021)Li, Gong, Liu, Wang, Qiao, and Cheng]{li2021metasaug}
Shuang Li, Kaixiong Gong, Chi~Harold Liu, Yulin Wang, Feng Qiao, and Xinjing Cheng.
\newblock Metasaug: Meta semantic augmentation for long-tailed visual recognition.
\newblock In \emph{Proceedings of the IEEE/CVF Conference on Computer Vision and Pattern Recognition}, pp.\  5212--5221, 2021.

\bibitem[Li et~al.(2022)Li, Dai, Ge, Liu, Shan, and Duan]{li2022uncertainty}
Xiaotong Li, Yongxing Dai, Yixiao Ge, Jun Liu, Ying Shan, and Ling-Yu Duan.
\newblock Uncertainty modeling for out-of-distribution generalization.
\newblock In \emph{ICLR}, 2022.

\bibitem[Li et~al.(2018{\natexlab{c}})Li, Tian, Gong, Liu, Liu, Zhang, and Tao]{li2018deep}
Ya~Li, Xinmei Tian, Mingming Gong, Yajing Liu, Tongliang Liu, Kun Zhang, and Dacheng Tao.
\newblock Deep domain generalization via conditional invariant adversarial networks.
\newblock In \emph{ECCV}, pp.\  624--639, 2018{\natexlab{c}}.

\bibitem[Li et~al.(2016)Li, Wang, Shi, Liu, and Hou]{li2016revisiting}
Yanghao Li, Naiyan Wang, Jianping Shi, Jiaying Liu, and Xiaodi Hou.
\newblock Revisiting batch normalization for practical domain adaptation.
\newblock \emph{arXiv preprint arXiv:1603.04779}, 2016.

\bibitem[Li et~al.(2019{\natexlab{b}})Li, Yang, Zhou, and Hospedales]{li2019feature}
Yiying Li, Yongxin Yang, Wei Zhou, and Timothy Hospedales.
\newblock Feature-critic networks for heterogeneous domain generalization.
\newblock In \emph{ICML}, pp.\  3915--3924, 2019{\natexlab{b}}.

\bibitem[Li \& Hoiem(2017)Li and Hoiem]{li2017learning}
Zhizhong Li and Derek Hoiem.
\newblock Learning without forgetting.
\newblock \emph{T-PAMI}, 40\penalty0 (12):\penalty0 2935--2947, 2017.

\bibitem[Liang et~al.(2020)Liang, Hu, and Feng]{liang2020we}
Jian Liang, Dapeng Hu, and Jiashi Feng.
\newblock Do we really need to access the source data? source hypothesis transfer for unsupervised domain adaptation.
\newblock In \emph{ICLR}, pp.\  6028--6039. PMLR, 2020.

\bibitem[Liu et~al.(2021{\natexlab{a}})Liu, Sun, Wang, Tang, Li, Qin, Chen, and Liu]{liu2021learning}
Chang Liu, Xinwei Sun, Jindong Wang, Haoyue Tang, Tao Li, Tao Qin, Wei Chen, and Tie-Yan Liu.
\newblock Learning causal semantic representation for out-of-distribution prediction.
\newblock \emph{NeurIPS}, 34, 2021{\natexlab{a}}.

\bibitem[Liu et~al.(2020)Liu, Dou, and Heng]{liu2020shape}
Quande Liu, Qi~Dou, and Pheng-Ann Heng.
\newblock Shape-aware meta-learning for generalizing prostate mri segmentation to unseen domains.
\newblock In \emph{MICCAI}, pp.\  475--485. Springer, 2020.

\bibitem[Liu et~al.(2021{\natexlab{b}})Liu, Lin, Cao, Hu, Wei, Zhang, Lin, and Guo]{liu2021swin}
Ze~Liu, Yutong Lin, Yue Cao, Han Hu, Yixuan Wei, Zheng Zhang, Stephen Lin, and Baining Guo.
\newblock Swin transformer: Hierarchical vision transformer using shifted windows.
\newblock In \emph{Proceedings of the IEEE/CVF international conference on computer vision}, pp.\  10012--10022, 2021{\natexlab{b}}.

\bibitem[Liu et~al.(2022)Liu, Mao, Wu, Feichtenhofer, Darrell, and Xie]{liu2022convnet}
Zhuang Liu, Hanzi Mao, Chao-Yuan Wu, Christoph Feichtenhofer, Trevor Darrell, and Saining Xie.
\newblock A convnet for the 2020s.
\newblock In \emph{Proceedings of the IEEE/CVF conference on computer vision and pattern recognition}, pp.\  11976--11986, 2022.

\bibitem[Long et~al.(2015)Long, Cao, Wang, and Jordan]{long2015learning}
Mingsheng Long, Yue Cao, Jianmin Wang, and Michael Jordan.
\newblock Learning transferable features with deep adaptation networks.
\newblock In \emph{ICML}, pp.\  97--105, 2015.

\bibitem[Mahajan et~al.(2021)Mahajan, Tople, and Sharma]{mahajan2021domain}
Divyat Mahajan, Shruti Tople, and Amit Sharma.
\newblock Domain generalization using causal matching.
\newblock In \emph{ICML}, pp.\  7313--7324, 2021.

\bibitem[Mao et~al.(2022)Mao, Chen, Jia, Zhang, Xue, and Li]{mao2022context}
Xiaofeng Mao, Yuefeng Chen, Xiaojun Jia, Rong Zhang, Hui Xue, and Zhao Li.
\newblock Context-aware robust fine-tuning.
\newblock \emph{arXiv preprint arXiv:2211.16175}, 2022.

\bibitem[Matsuura \& Harada(2020)Matsuura and Harada]{dg_mmld}
Toshihiko Matsuura and Tatsuya Harada.
\newblock Domain generalization using a mixture of multiple latent domains.
\newblock In \emph{AAAI}, 2020.

\bibitem[Park et~al.(2023)Park, Han, Kim, and Moon]{park2023test}
Jungwuk Park, Dong-Jun Han, Soyeong Kim, and Jaekyun Moon.
\newblock Test-time style shifting: Handling arbitrary styles in domain generalization.
\newblock \emph{arXiv preprint arXiv:2306.04911}, 2023.

\bibitem[Peng et~al.(2022)Peng, Lei, Hayat, Guo, and Li]{peng2022semantic}
Duo Peng, Yinjie Lei, Munawar Hayat, Yulan Guo, and Wen Li.
\newblock Semantic-aware domain generalized segmentation.
\newblock In \emph{CVPR}, 2022.

\bibitem[Peng et~al.(2019)Peng, Bai, Xia, Huang, Saenko, and Wang]{peng2019moment}
Xingchao Peng, Qinxun Bai, Xide Xia, Zijun Huang, Kate Saenko, and Bo~Wang.
\newblock Moment matching for multi-source domain adaptation.
\newblock In \emph{ICCV}, pp.\  1406--1415, 2019.

\bibitem[Piratla et~al.(2020)Piratla, Netrapalli, and Sarawagi]{piratla2020efficient}
Vihari Piratla, Praneeth Netrapalli, and Sunita Sarawagi.
\newblock Efficient domain generalization via common-specific low-rank decomposition.
\newblock In \emph{ICML}, pp.\  7728--7738, 2020.

\bibitem[Radford et~al.(2015)Radford, Metz, and Chintala]{radford2015unsupervised}
Alec Radford, Luke Metz, and Soumith Chintala.
\newblock Unsupervised representation learning with deep convolutional generative adversarial networks.
\newblock \emph{arXiv preprint arXiv:1511.06434}, 2015.

\bibitem[Radford et~al.(2021)Radford, Kim, Hallacy, Ramesh, Goh, Agarwal, Sastry, Askell, Mishkin, Clark, et~al.]{radford2021learning}
Alec Radford, Jong~Wook Kim, Chris Hallacy, Aditya Ramesh, Gabriel Goh, Sandhini Agarwal, Girish Sastry, Amanda Askell, Pamela Mishkin, Jack Clark, et~al.
\newblock Learning transferable visual models from natural language supervision.
\newblock In \emph{International conference on machine learning}, pp.\  8748--8763. PMLR, 2021.

\bibitem[Radosavovic et~al.(2020)Radosavovic, Kosaraju, Girshick, He, and Doll{\'a}r]{radosavovic2020designing}
Ilija Radosavovic, Raj~Prateek Kosaraju, Ross Girshick, Kaiming He, and Piotr Doll{\'a}r.
\newblock Designing network design spaces.
\newblock In \emph{Proceedings of the IEEE/CVF conference on computer vision and pattern recognition}, pp.\  10428--10436, 2020.

\bibitem[Rame et~al.(2022)Rame, Dancette, and Cord]{rame2022fishr}
Alexandre Rame, Corentin Dancette, and Matthieu Cord.
\newblock Fishr: Invariant gradient variances for out-of-distribution generalization.
\newblock In \emph{International Conference on Machine Learning}, pp.\  18347--18377. PMLR, 2022.

\bibitem[Rebuffi et~al.(2017)Rebuffi, Kolesnikov, Sperl, and Lampert]{rebuffi2017icarl}
Sylvestre-Alvise Rebuffi, Alexander Kolesnikov, Georg Sperl, and Christoph~H Lampert.
\newblock icarl: Incremental classifier and representation learning.
\newblock In \emph{CVPR}, pp.\  2001--2010, 2017.

\bibitem[Rojas-Carulla et~al.(2018)Rojas-Carulla, Sch{\"o}lkopf, Turner, and Peters]{rojas2018invariant}
Mateo Rojas-Carulla, Bernhard Sch{\"o}lkopf, Richard Turner, and Jonas Peters.
\newblock Invariant models for causal transfer learning.
\newblock \emph{JMLR}, 19\penalty0 (1):\penalty0 1309--1342, 2018.

\bibitem[Saito et~al.(2020)Saito, Kim, Sclaroff, and Saenko]{saito2020universal}
Kuniaki Saito, Donghyun Kim, Stan Sclaroff, and Kate Saenko.
\newblock Universal domain adaptation through self supervision.
\newblock In \emph{NeuIPS}, volume~33, pp.\  16282--16292, 2020.
\newblock URL \url{https://proceedings.neurips.cc/paper/2020/file/bb7946e7d85c81a9e69fee1cea4a087c-Paper.pdf}.

\bibitem[Schneider et~al.(2020)Schneider, Rusak, Eck, Bringmann, Brendel, and Bethge]{schneider2020improving}
Steffen Schneider, Evgenia Rusak, Luisa Eck, Oliver Bringmann, Wieland Brendel, and Matthias Bethge.
\newblock Improving robustness against common corruptions by covariate shift adaptation.
\newblock \emph{NeuIPS}, 33, 2020.

\bibitem[Shi et~al.(2021)Shi, Seely, Torr, Siddharth, Hannun, Usunier, and Synnaeve]{shi2021gradient}
Yuge Shi, Jeffrey Seely, Philip~HS Torr, N~Siddharth, Awni Hannun, Nicolas Usunier, and Gabriel Synnaeve.
\newblock Gradient matching for domain generalization.
\newblock \emph{arXiv preprint arXiv:2104.09937}, 2021.

\bibitem[Shu et~al.(2023)Shu, Guo, Wu, Wang, Wang, and Long]{shu2023clipood}
Yang Shu, Xingzhuo Guo, Jialong Wu, Ximei Wang, Jianmin Wang, and Mingsheng Long.
\newblock Clipood: Generalizing clip to out-of-distributions.
\newblock \emph{arXiv preprint arXiv:2302.00864}, 2023.

\bibitem[Singh et~al.(2022)Singh, Gustafson, Adcock, de~Freitas~Reis, Gedik, Kosaraju, Mahajan, Girshick, Doll{\'a}r, and Van Der~Maaten]{singh2022revisiting}
Mannat Singh, Laura Gustafson, Aaron Adcock, Vinicius de~Freitas~Reis, Bugra Gedik, Raj~Prateek Kosaraju, Dhruv Mahajan, Ross Girshick, Piotr Doll{\'a}r, and Laurens Van Der~Maaten.
\newblock Revisiting weakly supervised pre-training of visual perception models.
\newblock In \emph{Proceedings of the IEEE/CVF Conference on Computer Vision and Pattern Recognition}, pp.\  804--814, 2022.

\bibitem[Sun \& Saenko(2016)Sun and Saenko]{sun2016deep}
Baochen Sun and Kate Saenko.
\newblock Deep coral: Correlation alignment for deep domain adaptation.
\newblock In \emph{ECCV}, pp.\  443--450, 2016.

\bibitem[Sun et~al.(2021)Sun, Wu, Zheng, Liu, Chen, Qin, and Liu]{sun2021recovering}
Xinwei Sun, Botong Wu, Xiangyu Zheng, Chang Liu, Wei Chen, Tao Qin, and Tie-Yan Liu.
\newblock Recovering latent causal factor for generalization to distributional shifts.
\newblock \emph{NeurIPS}, 34, 2021.

\bibitem[Sun et~al.(2020)Sun, Wang, Liu, Miller, Efros, and Hardt]{sun2020test}
Yu~Sun, Xiaolong Wang, Zhuang Liu, John Miller, Alexei Efros, and Moritz Hardt.
\newblock Test-time training with self-supervision for generalization under distribution shifts.
\newblock In \emph{ICML}, pp.\  9229--9248. PMLR, 2020.

\bibitem[Tan \& Le(2021)Tan and Le]{tan2021efficientnetv2}
Mingxing Tan and Quoc Le.
\newblock Efficientnetv2: Smaller models and faster training.
\newblock In \emph{International conference on machine learning}, pp.\  10096--10106. PMLR, 2021.

\bibitem[Tolstikhin et~al.(2021)Tolstikhin, Houlsby, Kolesnikov, Beyer, Zhai, Unterthiner, Yung, Steiner, Keysers, Uszkoreit, et~al.]{tolstikhin2021mlp}
Ilya~O Tolstikhin, Neil Houlsby, Alexander Kolesnikov, Lucas Beyer, Xiaohua Zhai, Thomas Unterthiner, Jessica Yung, Andreas Steiner, Daniel Keysers, Jakob Uszkoreit, et~al.
\newblock Mlp-mixer: An all-mlp architecture for vision.
\newblock \emph{Advances in neural information processing systems}, 34:\penalty0 24261--24272, 2021.

\bibitem[Torralba \& Efros(2011)Torralba and Efros]{torralba2011unbiased}
Antonio Torralba and Alexei~A Efros.
\newblock Unbiased look at dataset bias.
\newblock In \emph{CVPR}, pp.\  1521--1528, 2011.

\bibitem[Touvron et~al.(2021)Touvron, Cord, Douze, Massa, Sablayrolles, and J{\'e}gou]{touvron2021training}
Hugo Touvron, Matthieu Cord, Matthijs Douze, Francisco Massa, Alexandre Sablayrolles, and Herv{\'e} J{\'e}gou.
\newblock Training data-efficient image transformers \& distillation through attention.
\newblock In \emph{International conference on machine learning}, pp.\  10347--10357. PMLR, 2021.

\bibitem[Vapnik(1991)]{vapnik1991principles}
Vladimir Vapnik.
\newblock Principles of risk minimization for learning theory.
\newblock \emph{Advances in neural information processing systems}, 4, 1991.

\bibitem[Vapnik(1998)]{vapnik1998statistical}
Vladimir Vapnik.
\newblock \emph{Statistical learning theory. 1998}, volume~3.
\newblock Wiley, New York, 1998.

\bibitem[Venkateswara et~al.(2017)Venkateswara, Eusebio, Chakraborty, and Panchanathan]{venkateswara2017deep}
Hemanth Venkateswara, Jose Eusebio, Shayok Chakraborty, and Sethuraman Panchanathan.
\newblock Deep hashing network for unsupervised domain adaptation.
\newblock In \emph{CVPR}, pp.\  5018--5027, 2017.

\bibitem[Wang et~al.(2021)Wang, Shelhamer, Liu, Olshausen, and Darrell]{wang2020tent}
Dequan Wang, Evan Shelhamer, Shaoteng Liu, Bruno Olshausen, and Trevor Darrell.
\newblock Tent: Fully test-time adaptation by entropy minimization.
\newblock In \emph{ICLR}, 2021.

\bibitem[Wang et~al.(2022{\natexlab{a}})Wang, Lan, Liu, Ouyang, Qin, Lu, Chen, Zeng, and Yu]{wang2022generalizing}
Jindong Wang, Cuiling Lan, Chang Liu, Yidong Ouyang, Tao Qin, Wang Lu, Yiqiang Chen, Wenjun Zeng, and Philip Yu.
\newblock Generalizing to unseen domains: A survey on domain generalization.
\newblock \emph{TKDE}, 2022{\natexlab{a}}.

\bibitem[Wang et~al.(2022{\natexlab{b}})Wang, Fink, Van~Gool, and Dai]{wang2022continual}
Qin Wang, Olga Fink, Luc Van~Gool, and Dengxin Dai.
\newblock Continual test-time domain adaptation.
\newblock In \emph{Proceedings of the IEEE/CVF Conference on Computer Vision and Pattern Recognition}, pp.\  7201--7211, 2022{\natexlab{b}}.

\bibitem[Wang et~al.(2019)Wang, Jin, Long, Wang, and Jordan]{wang2019transferable}
Ximei Wang, Ying Jin, Mingsheng Long, Jianmin Wang, and Michael~I Jordan.
\newblock Transferable normalization: Towards improving transferability of deep neural networks.
\newblock In \emph{NeurIPS}, pp.\  1951--1961, 2019.

\bibitem[Wu et~al.(2018)Wu, Xiong, Yu, and Lin]{wu2018unsupervised}
Zhirong Wu, Yuanjun Xiong, Stella~X Yu, and Dahua Lin.
\newblock Unsupervised feature learning via non-parametric instance discrimination.
\newblock In \emph{Proceedings of the IEEE conference on computer vision and pattern recognition}, pp.\  3733--3742, 2018.

\bibitem[Yeh et~al.(2021)Yeh, Yang, Yuen, and Harada]{yeh2021sofa}
Hao-Wei Yeh, Baoyao Yang, Pong~C Yuen, and Tatsuya Harada.
\newblock Sofa: Source-data-free feature alignment for unsupervised domain adaptation.
\newblock In \emph{WACV}, pp.\  474--483, 2021.

\bibitem[You et~al.(2021)You, Li, and Zhao]{you2021test}
Fuming You, Jingjing Li, and Zhou Zhao.
\newblock Test-time batch statistics calibration for covariate shift.
\newblock \emph{arXiv preprint arXiv:2110.04065}, 2021.

\bibitem[Zenke et~al.(2017)Zenke, Poole, and Ganguli]{zenke2017continual}
Friedemann Zenke, Ben Poole, and Surya Ganguli.
\newblock Continual learning through synaptic intelligence.
\newblock In \emph{International Conference on Machine Learning}, pp.\  3987--3995. PMLR, 2017.

\bibitem[Zhang et~al.(2023{\natexlab{a}})Zhang, Qi, Shi, and Gao]{zhang2023domainadaptor}
Jian Zhang, Lei Qi, Yinghuan Shi, and Yang Gao.
\newblock Domainadaptor: A novel approach to test-time adaptation.
\newblock \emph{arXiv preprint arXiv:2308.10297}, 2023{\natexlab{a}}.

\bibitem[Zhang et~al.(2021)Zhang, Gu, Matsuo, and Iwasawa]{zhang2021domain}
Xin Zhang, Shixiang~Shane Gu, Yutaka Matsuo, and Yusuke Iwasawa.
\newblock Domain prompt learning for efficiently adapting clip to unseen domains.
\newblock \emph{arXiv e-prints}, pp.\  arXiv--2111, 2021.

\bibitem[Zhang et~al.(2023{\natexlab{b}})Zhang, Wang, Jin, Yuan, Zhang, Wang, Jin, and Tan]{zhang2023adanpc}
Yifan Zhang, Xue Wang, Kexin Jin, Kun Yuan, Zhang Zhang, Liang Wang, Rong Jin, and Tieniu Tan.
\newblock Adanpc: Exploring non-parametric classifier for test-time adaptation.
\newblock In \emph{International Conference on Machine Learning}, pp.\  41647--41676. PMLR, 2023{\natexlab{b}}.

\bibitem[Zhou \& Levine(2021)Zhou and Levine]{zhou2021training}
Aurick Zhou and Sergey Levine.
\newblock Training on test data with bayesian adaptation for covariate shift.
\newblock \emph{arXiv preprint arXiv:2109.12746}, 2021.

\bibitem[Zhou et~al.(2021)Zhou, Yang, Qiao, and Xiang]{zhou2021domain}
Kaiyang Zhou, Yongxin Yang, Yu~Qiao, and Tao Xiang.
\newblock Domain generalization with mixstyle.
\newblock In \emph{ICLR}, 2021.

\bibitem[Zhou et~al.(2022)Zhou, Liu, Qiao, Xiang, and Loy]{zhou2022domain}
Kaiyang Zhou, Ziwei Liu, Yu~Qiao, Tao Xiang, and Chen~Change Loy.
\newblock Domain generalization: A survey.
\newblock \emph{TPAMI}, 2022.

\end{thebibliography}
\bibliographystyle{iclr2024_conference}
\clearpage
\appendix
{\Large\textbf{Appendix}}
\section{Summary}
This appendix describes more details of the ICLR 2024 submission, titled \textit{Towards Unified and Efficient Domain Generalization}. The appendix is organized as follows:
\begin{itemize}
	\item \S~\ref{sec:theory} theoretically discusses why \Model can perform better under the setting of test-time adaptation than the BN-based approaches.
	\item \S~\ref{sec:imple} provides more implementation details and analytical experiment for the hyper-parameter $\sigma$ of Equation~\ref{eq:margin}.
	\item \S~\ref{sec:pseu} summarizes the pseudo-code of our proposed \Model.
	\item \S~\ref{sec:vis} illustrates the effectiveness of the proposed \Model by quantities of T-SNE visualization results compared with existing advanced methods including TENT~\cite{wang2020tent} and TAST~\cite{jang2022test}.
	\item \S~\ref{sec:supp:exp} exhibits a more detailed comparison between the other test-time schemes and provides improvements of each visual backbone on all domains.
	
\end{itemize}

\section{Theoretical Insight}~\label{sec:theory}
To mitigate catastrophic forgetting during test-time adaptation, existing methods such as TENT~\cite{wang2020tent}, SHOT~\cite{liang2020we}, and TAST~\cite{jang2022test} propose adapting the parameters of Batch Normalization (\texttt{BN}) layers. We argue that adapting only the \texttt{BN} layers may be insufficient for effectively handling the unseen novel domains compared to adapting the entire network parameters. To substantiate our claim, we provide a theoretical analysis from two perspectives:

1) Neural Tangent Kernels (NTK)~\cite{jacot2018neural}: NTK is a kernel that elucidates the evolution of neural networks during training via gradient descent. It bridges the gap between neural networks and classical kernel methods in machine learning. When the width of the hidden layers in a neural network approaches infinity, the network's training behavior becomes more predictable. In this paper, we employ neural tangent kernels to assess the network's ability to generalize to unseen domains $\bm{x}^T\in \mathcal{T}$ while training on the source domain $\bm{x}^S\in \mathcal{S}$.

2) Gradient descent process of \texttt{BN} layers restricts the expansion of neural tangent kernels: We argue that solely adapting the \texttt{BN} layers could limit the growth of neural tangent kernels, affecting the model's generalization capability for unseen domains.

In summary, our theoretical analysis highlights the potential limitations of adapting only the \texttt{BN} layers in handling novel domains and suggests that a more comprehensive approach might be necessary to achieve better generalization.\par
\subsection{Neural Tangent Kernel in DG}~\label{sec:theory:nek+dg}
For domain generalization, the networks can be formulated from two parts: feature extractor $f(\cdot)$ and classifier $q(\cdot)$. And Neural Tangent Kernel $K_{\cdot,\cdot}$ formulates the impact of gradients between different instances $x^S, x^T$ to the learning representations of the neural network in the gradient decent progress:
\begin{equation}
	\small
	K_{x^S, x^{T}} = k(x^{S}, x^{T}) =\lim _{\eta \rightarrow 0} \frac{f(x, \theta+\eta[\frac{\nabla f_{\theta}(x)}{\nabla \theta}]-f(x, \theta) )}{\eta}, 
	\label{eq:ntk}
\end{equation}
where $\eta$ is the learning rate and $\theta$ denotes the parameters.
Accordingly, we can obtain the relationship between parameters and learning representations with learning rate $\eta$ and neural tangent kernel $K_{x^{s} x^{T}}$ on the source and target domains:
\begin{equation}
	\small
	\begin{aligned}
		f( x^{\top}, \theta + \eta [\frac{\nabla f(x^{T})}{\eta \theta}])& =k(x^{S}, x^{T}) \eta + f(x^{S}, \theta) -f(x^{T}, \theta) = K_{x^{S} x^{T}} \cdot \eta+f(x^{S}, \theta)-f(x^{T}, \theta).
	\end{aligned}
	\label{eq:f+k}
\end{equation}
Meanwhile, we can also obtain the formula of neural tangent kernels according to its definition~\cite{jacot2018neural}:
\begin{equation}
	\begin{aligned}
		K_{x^{S} x^{T}}&= \mathbb{E}_{\theta \sim \omega}[f(x^{S}, \theta) \cdot f(x^{T}, \theta)] \hspace{1.5mm}= \mathbb{E}_{\theta \sim \omega}[\langle\frac{\partial f(x^{S}, \theta)}{\partial \theta}, \frac{\partial f(x^{T}, \theta)}{\partial \theta}\rangle],
	\end{aligned}
	\label{eq:kernel}
\end{equation}
where $\mathbb{E}_{\theta \sim \omega}(\cdot)$ indicates the mathematical expectation of network parameter in the parameters space $\omega$, and $\frac{\partial f(x^{S}, \theta)}{\partial \theta}$ and $\frac{\partial f(x^{T}, \theta)}{\partial \theta}$ denotes the gradients of the parameters with representation on the source and target domains, respectively. And the neural tangent kernels in domain generalization can be regarded as the inner product of the gradients for learning representations on the source and target domains.
\subsection{Backward Gradients in BN Layers}~\label{sec:theory:bn}
Batch Normalization operation is focused on introducing means $\mu$ and standard error $\sigma^2$ to normalize learning representations in a mini-batch $\mathcal{B}=\{x_{1}, x_{2}, \ldots , x_{m}\}$ containing $m$ samples. And \texttt{BN} layer introduces linear projection with two learnable parameters $\gamma$ and $\beta$ based on the Batch Normalization operation. Its computation can be formulated as:
\begin{equation}
	\begin{aligned}
		&\mu_{\mathcal{B}} \leftarrow \frac{1}{m} \sum_{i=1}^{m} x_{i}, \hspace{1.0mm}
		\sigma_{\mathcal{B}}^{2} \leftarrow \frac{1}{m} \sum_{i=1}^{m}(x_{i}-\mu_{B})^{2}, \hspace{1.5mm} \hat{x}_{i} \leftarrow \frac{x_{i}- \mu_{\beta}}{\sqrt{\sigma_{\beta}^{2}+\varepsilon}},
		\hspace{1.0mm} \hat{y}_i \leftarrow \gamma \hat{x}_i + \beta,
	\end{aligned}
	\label{eq:bn}
\end{equation}
where $\hat{y}_i$ is the output of \texttt{BN} layer, which can be simplified as: $y_{i}= {\rm BN}_{\gamma, \beta}(x_{i})$, $\varepsilon$ is a value to smooth the computation. And we can obtain the formula to describe it with involved variables including $m,\gamma,\beta,\sigma,{x}_i$:
\begin{equation}
	\hat{y}_{i}= \gamma \frac{x_{i}-\frac{1}{m} \sum_{i=1}^{m} x_{i}}{\sqrt{\frac{1}{m} \sum_{i=1}^{m}(x_{i}-\frac{1}{m} \sum_{i=1}^{m} x_{i})^{2}+\varepsilon}}.
	\label{eq:y}
\end{equation} 
According to Equation~\ref{eq:y}, we can utilize chain rules to calculate the gradients between representations and input with intermediate variables:
\begin{equation}
	\small
	\begin{aligned}
		& \frac{\partial \hat{x}_{i}}{\partial \mu}= - \frac{1}{\sqrt{\sigma^{2}+\varepsilon}}, \hspace{1.5mm}\frac{\partial \hat{x}}{\partial \sigma^{2}}=\sum_{i=1}^{m}(x_{i}-\mu)^{-\frac{1}{2}}, (\sigma^{2}+\epsilon)^{-\frac{3}{2}}, \hspace{1.5mm}\frac{\partial \sigma^{2}}{\partial \mu}=\frac{1}{m} \sum_{i=1}^{m} - (x_{i}-\mu).
	\end{aligned}
	\label{eq:grad:mu+sigma}
\end{equation}
Furthermore, we can derive the gradient of learning representation $f$ in \texttt{BN} layers according to Equation~\ref{eq:bn} for learnable parameters $\gamma, \beta$:
\begin{equation}
	\begin{aligned}
		&\frac{\partial f}{\partial \gamma}=\frac{\partial f}{\partial y_{i}} \cdot \frac{\partial y_{i}}{\partial \gamma}=\sum_{i=1}^{m} \frac{\partial f}{\partial y_{i}} \cdot \hat{x}, \hspace{2.5mm}\frac{\partial f}{\partial \beta}=\frac{\partial f}{\partial y_{i}} \cdot \frac{\partial y}{\partial \beta}=\sum_{i=1}^{m} \frac{\partial f}{\partial y_{i}}. \\
	\end{aligned}
\end{equation} 
And we can also obtain the gradients for representations and statistic variables in the mini-batch by chain rules:
\begin{equation}
	\begin{aligned}
		& \frac{\partial f}{\partial \hat{x}_{i}}=\frac{\partial f}{\partial y} \cdot \frac{\partial y_{i}}{\partial x_{i}}=\frac{\partial f}{\partial y_{i}} \cdot \gamma, \hspace{2.5mm} \frac{\partial f}{\partial \sigma^{2}}=\frac{\partial f}{\partial \hat{x}} \cdot \frac{\partial \hat{x}}{\partial \sigma^{2}}, \hspace{1.5mm}\frac{\partial f}{\partial u}=\frac{\partial f}{\partial \hat{x}_{i}}  \frac{\partial \hat{x}_{i}}{\partial \mu} + \frac{\partial f}{\partial \sigma^{2}} \frac{\partial \sigma^{2}}{\partial \mu},
	\end{aligned}
	\label{eq:grad:f+mu+sigma}
\end{equation} 
and we have obtained $\frac{\partial \hat{x}_{i}}{\partial \mu}$ and $\frac{\partial \hat{x}}{\partial \sigma^{2}}$ in Equation~\ref{eq:grad:mu+sigma}. Then, for the unknown gradients of $\frac{\partial f}{\partial \mu}$, we can utilize the results in Equation~\ref{eq:grad:mu+sigma} to simplify:
\begin{equation}
	\small
	\begin{aligned}
		& \frac{\partial f}{\partial \mu}=(\sum_{i=1}^{m} \frac{\partial f}{\partial \hat{x}_i} \cdot \frac{-1}{\sqrt{\sigma^{2}+\varepsilon}})+(\frac{\partial f}{\partial \sigma^{2}} \cdot \frac{1}{m} \sum_{i=1}^{m} -2(x_{i}- \mu)), \\
		& =\left(\sum_{i=1}^{m} \frac{\partial f}{\partial \hat{x}_{i}}-\frac{-1}{\sqrt{\sigma^{2}+\varepsilon}}\right)+\left(\frac{\partial f}{\partial \sigma^{2}}  \cdot\left(\frac{-2}{m} \sum_{i=1}^{m} x_{i}+\frac{2}{m} \sum_{i=1}^{m} u\right)\right), \\
		& =\left(\sum_{i=1}^{m} \frac{\partial f}{\partial \hat{x}_{i}} \cdot \frac{-1}{\sqrt{\sigma^{2}+\varepsilon}}\right)+\left(\frac{\partial f}{\partial \sigma^{2}} \cdot\left(-2u+\frac{2m \cdot u}{m}\right)\right) =\sum_{i=1}^{m} \frac{\partial f}{\partial \hat{x}_{i}} \cdot \frac{-1}{\sqrt{\sigma^{2}+\varepsilon}},
	\end{aligned}
	\label{eq:grad:f+mu}
\end{equation} 
In addition, the gradients of $\frac{\partial f}{\partial x_{i}}$ is directly calculated with chain rules as:$\frac{\partial f}{\partial x_{i}}=\frac{\partial f}{\partial x_{i}} \cdot \frac{\partial \hat{x}_{i}}{\partial x_{i}}+\frac{\partial f}{\partial u}+\frac{\partial u}{\partial x_{i}}+\frac{\partial f}{\partial \sigma^{2}} \cdot \frac{\partial \sigma^{2}}{\partial x_{i}}.$ Referring to Equation~\ref{eq:bn}, these gradients can be simply obtained $\frac{\partial \hat{x}_{i}}{\partial x_{i}}=\frac{1}{\sqrt{\sigma^{2}+\varepsilon}}$, $\frac{\partial u}{\partial x_{i}}=\frac{1}{m}$, and $\frac{\partial \sigma^{2}}{\partial x_{i}}=\frac{2\left(x_{i}-u\right)}{m}$. Therefore, we can compute the gradient $\frac{\partial f}{\partial x_{i}}$ as:
\begin{equation}
	\small
	\frac{\partial f}{\partial x_{i}}=\left(\frac{\partial f}{\partial \hat{x}_{i}} \cdot \frac{1}{\sqrt{\sigma^{2}+\varepsilon}}\right)+\left(\frac{\partial f}{\partial u} \cdot \frac{1}{m}\right)+\left(\frac{\partial f}{\partial \sigma^{2}} \cdot \frac{2\left(x_{i}-u\right)}{m}\right),
	\label{eq:grad:f+x}
\end{equation}
According to Equation~\ref{eq:grad:mu+sigma}, \ref{eq:grad:f+mu+sigma}, \ref{eq:grad:f+mu}, we compute the gradient as:
\begin{equation}
	\small
	\begin{aligned}
		& \frac{\partial f}{\partial x_{i}}=\left(\frac{\partial f}{\partial \hat{x}_{i}} \cdot \frac{1}{\sqrt{\sigma^{2}+\varepsilon}}\right)+\left(\frac{\partial f}{\partial u} \cdot \frac{1}{m}\right)+\left(\frac{\partial f}{\partial \sigma^{2}} \cdot \frac{2\left(x_{i}-u \right)}{m}\right) \\
		& =\left(\frac{\partial f}{\partial \hat{x}_{i}} \cdot \frac{1}{\sqrt{\sigma^{2}+\varepsilon}}\right)+\left(\frac{1}{m} \sum_{j=1}^{m} \frac{\partial f}{\partial \hat{x}_{j}} \cdot \frac{-1}{\sqrt{\sigma^{2}+\varepsilon}}\right) -\left(\frac{1}{2} \sum_{j=1}^{m} \frac { \partial f } {\partial \hat{x}} \left(x_{j}-u \right) \cdot\left(\sigma^{2}+\varepsilon\right)^{-\frac{3}{2}} \right. \cdot \left.\frac{2\left(x_{i}-u \right)}{m}\right) \\
		& =\left(\frac{\partial f}{\partial \hat{x}} \cdot\left(\sigma^{2}+\varepsilon\right)^{-\frac{1}{2}}\right)-\left(\frac{\left(\sigma^{2}+\varepsilon\right)^{-\frac{1}{2}}}{m} \cdot \sum_{j=1}^{m} \frac{\partial f}{\partial \hat{x}_j}\right) +\left(\frac{\left(\sigma^{2}+\varepsilon\right)^{-\frac{1}{2}}}{m} \cdot \hat{x_{i}} \cdot \sum_{j=1}^{m} \frac{\partial f}{\partial \hat{x}_j} \cdot \hat{x_{j}}\right) \\
		& =\frac{\left(\sigma^{2}+\varepsilon\right)^{-\frac{1}{2}}}{m}\left[m \frac{\partial f}{\partial \hat{x}_{i}}-\sum_{j=1}^{m} \frac{\partial f}{\partial \widehat{x_{j}}}-\hat{x}_i \sum_{\hat{v}=1}^{m} \frac{\partial f}{\partial \hat{x}_{j}} \cdot \widehat{x_{j}}\right] \\
	\end{aligned}
	\label{eq:grad:f+xi+simple}
\end{equation}
Therefore, refer to Equation~\ref{eq:grad:f+mu+sigma}, we can obtain the gradients $\frac{\partial f}{\partial \hat{x}_{i}}$ related to $\frac{\partial f}{\partial y_i}$, which can be directly calculated with known gradients including $\frac{\partial f}{\partial \hat{x}_{i}}=\frac{\partial f}{\partial y_{i}} \cdot \gamma, \ \  \frac{\partial f}{\partial \beta}=\sum_{i=1}^{m} \frac{\partial f}{\partial y_{i}}, \ \ \frac{\partial f}{\partial \gamma}=\sum_{i=1}^{m} \frac{\partial f}{\partial y_{i}} \cdot \hat{x}_{i}$. Finally, the backward gradients in the BN layer can be computed as:
\begin{equation}
	\frac{\partial f}{\partial x_{i}}=\frac{m\frac{\partial f}{\partial x_{i}}-\sum_{j=1}^{m} \frac{\partial f}{\partial \hat{x}_{j}}- \hat{x}_{i} \sum_{j=1}^{m} \frac{\partial f}{\partial \hat{x}_{j}} \cdot x_{j}}{m \sqrt{\sigma^{2}+\varepsilon}}
	\label{eq:grad+bn}
\end{equation}
According to results in Equation~\ref{eq:ntk} and \ref{eq:grad+bn}, we can describe the adaptation process for these approaches adapting \texttt{BN} layers:
\begin{equation}
	\small
	\begin{aligned}
		&  K_{x^{S},x^{T}}=E_{\theta \sim W}\left[<\frac{\partial f\left(x^{S}, \theta\right)}{\partial \theta}, \frac{\partial f\left(x^{T}, \theta \right)}{\partial \theta}>\right] \\
		& \frac{\partial \mathtt{BN}(x)}{\partial x_i} = \frac{m\frac{\partial f }{\partial \hat{x}_i } - \sum_{j=1}^{m}\frac{\partial f}{\partial \hat{x}_j} - \hat{x}_i \sum_{j=1}^{m} \frac{\partial f}{\partial \hat{x}_j}  \hat{x}_j }{m \sqrt{\sigma^2 + \varepsilon}} \\
		&  K_{x^{S},x^{T}}^{\mathtt{BN}}=\frac{1}{m}\sum_{i = 1}^m[(\frac{\partial \mathtt{BN}(x^T)}{\partial x^T_i}\cdot\frac{\partial \mathtt{BN}(x^S)}{\partial x^S_i})/(\|\frac{\partial \mathtt{BN}(x^T)}{\partial x^T_i}\|\cdot\|\frac{\partial \mathtt{BN}(x^S)}{\partial x^S_i}\|)] \\
		& =  \frac{1}{m}\sum_{i = 1}^m[\frac{m\frac{\partial f }{\partial \hat{x}^T_i } - \sum_{j=1}^{m}\frac{\partial f}{\partial \hat{x}^T_j} - \hat{x}^T_i \sum_{j=1}^{m} \frac{\partial f}{\partial \hat{x}^T_j}  \hat{x}^T_j }{m \sqrt{\sigma^2 + \varepsilon}}] \cdot [\frac{m\frac{\partial f }{\partial \hat{x}^S_i } - \sum_{j=1}^{m}\frac{\partial f}{\partial \hat{x}^S_j} - \hat{x}^S_i \sum_{j=1}^{m} \frac{\partial f}{\partial \hat{x}^S_j}  \hat{x}^S_j }{m \sqrt{\sigma^2 + \varepsilon}}] \\
		& \hspace{20mm}/(\|\frac{\partial \mathtt{BN}(x^T)}{\partial x^T_i}\|\cdot\|\frac{\partial \mathtt{BN}(x^S)}{\partial x^S_i}\|) \\
		& \leq \frac{1}{(\sigma^2+\varepsilon)}[\frac{\partial f }{\partial \hat{x}^S_i} -\frac{\partial f }{\partial \hat{x}^S_j}-\hat{x_{i}^S}\cdot \frac{\partial f }{\partial \hat{x}^S_j}\hat{x_{j}^S} ]^2 / (\|\frac{\partial \mathtt{BN}(x^S)}{\partial x^S_i}\|^2) \leq K_{x^{S},x^{S}}
	\end{aligned}
	\label{eq:ntk+bn}
\end{equation}
\subsection{Conclusion}
Referring to Equation~\ref{eq:ntk+bn}, it illustrates that \textbf{neural tangent kernels for networks pretrained on source has less width to get transferred on target domains if the algorithm is focused on adapting parameters of the \texttt{BN} layers.} Therefore, we propose to adapt the comprehensive parameter space to enhance the ability to generalize on the novel environments.

\section{Implementation Details}~\label{sec:imple}

We split the dataset into training and validation sets, where 80\% of the samples are randomly selected as the training data and the remaining 20\% is the validation data. The validation set is exploited to select the hyper-parameter with the criterion of maximizing the accuracy of the validation set. 

In the test-time adaptation setting, the adaptation starts with a trained source model, which will be trained using source data in advance. To build a test-time adaptation method for domain generalization (DG), we first acquired a source model with data from multiple source domains. In addition, two pre-training algorithms are exploited for obtaining the source models, i.e., ERM and CORAL. The experiment results are the averages of three trials with different random seeds. We train the source model using an Adam optimizer with a learning rate of $5e^{-5}$ and a batch size of 32. The learning rate for adaptation phase is varying $\in \{5e^{-4}, 5e^{-5}, 5e^{-6}\}$.

\begin{figure}[t]
	\centering
	\includegraphics[width = 0.83\linewidth]{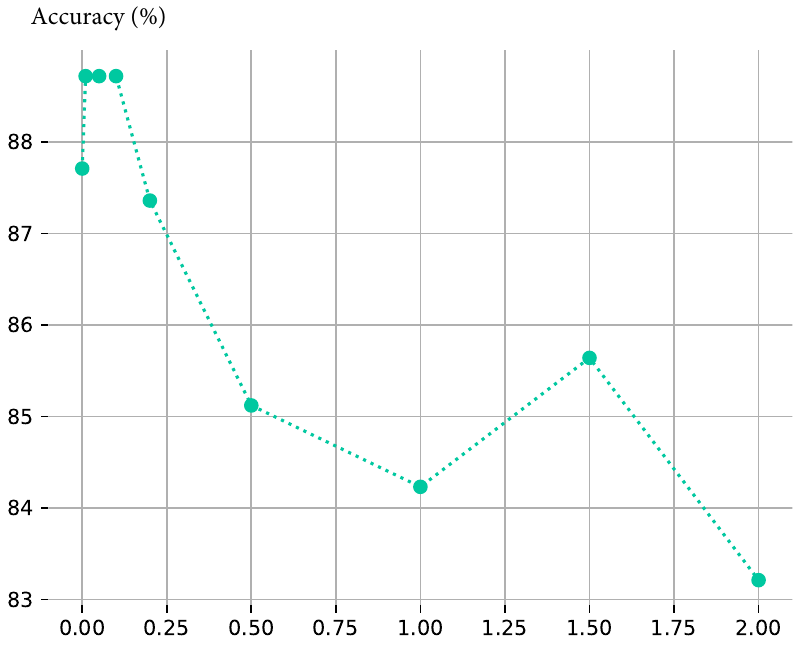}
	\caption{Hyper-parameter impact of $\sigma$.}
	\label{fig:hyer}
\end{figure}

\begin{figure*}[t]
	\centering
	\includegraphics[width = 0.96\linewidth]{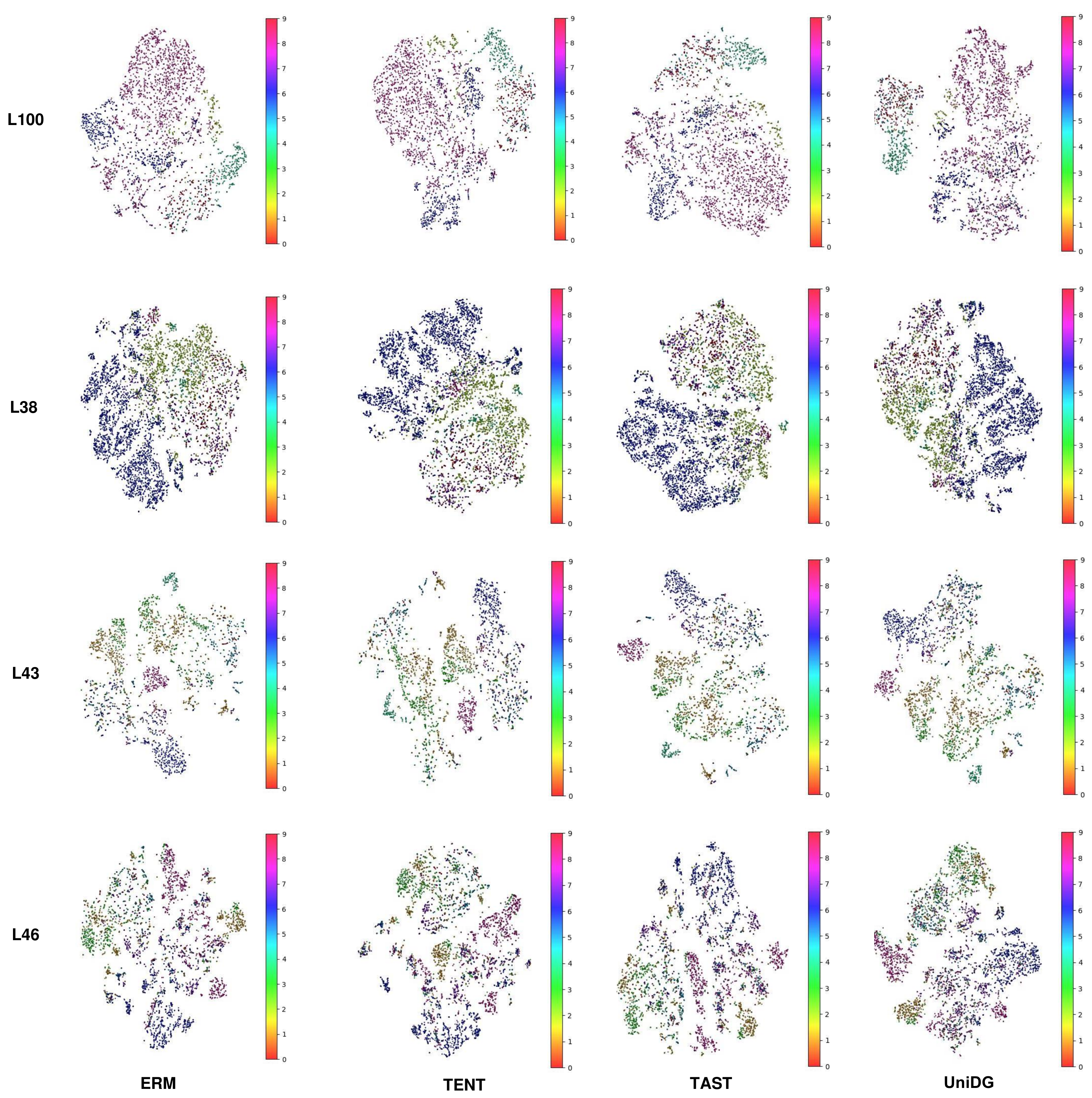}
	\caption{Qualitative Results of \Model on challenging TerraIncognita~\cite{beery2018recognition} dataset. \Model (4th column) aggregates intra-class feature embeddings better with the pretrained ERM model (1st column) compared with advanced methods including TENT~\cite{wang2020tent} (2nd column), and TAST~\cite{jang2022test} (3rd column).}
	\label{fig:vis-tsne}
\end{figure*}
In Figure~\ref{fig:hyer}, we visualize the sensitivity of hyper-parameter $\sigma$ in Equation~\ref{eq:margin} on the PACS~\cite{torralba2011unbiased} dataset. We take hyper-parameter $\sigma\in\{ 0, 0.01, 0.05, 0.10, 0.15, 0.20, 0.50, 1.0, 1.5, 2.0\}.$ It is also worth noting that when $\sigma$ is set in $[0.01,0.05,0.15,0.20]$, the network achieves better performances on the PACS dataset. Therefore, we choose to set $\sigma=0.15$.
\section{Pseudo code for \Model} ~\label{sec:pseu}
For feature extractor $f$, we detach the gradient and freeze the parameters of the trained source model, and utilize it to extract the representation of images from target domains as the source knowledge. And we initialize another new network $\theta^\prime$ with the trained parameters of $\theta$ and formulate the representation discrepancy between $f(\bm{x})$ and $f^\prime(\bm{x})$ via Frobenius norm $\|\cdot\|_F$:
\begin{equation}
	\begin{aligned}
		{{\cal L}_{m}} &= \frac{1}{\|D_T\|}\sum_{i = 1}^{\|D_T\|} {{{[\|{f^\prime(\bm{x}_i)}} - {f(\bm{x}_i)}\|_F^2  - \sigma ]} }.
	\end{aligned}
\end{equation}
For representations $f^\prime(\bm{x})$, we utilize a linear layer to work as a classifier and obtain the classification results $q^\prime(f^\prime(\bm{x}))$. Then we take the $\mathtt{softmax}$ entropy as the loss function to update the classifier:
\begin{equation}
	{{\cal L}_{e}} = - \frac{1}{N_b}\sum\nolimits_{i = 1}^{N_b} \log [\frac{\exp (q^\prime(f^\prime(\bm{x}_i)))}{\sum \exp q^\prime(f^\prime(\bm{x}))}].
	\label{eq:ent}
\end{equation}
The memory bank is set to store the class-wise prototypes of each class $\mathcal{M}=\bigcup_{i=1}^c\{\bm{p}_i\}, \bm{p}_i\in \mathbb{R}^{C}$, where $\mathcal{M}$, $c$, and $C$ denote the memory bank, the number of classes, and feature dimensions. For each class $i$, the prototype $\bm{p}_i$ is initialized with parameters of classifier $\bm{p}_iarrow q^\prime(\cdot)_i$:
\begin{equation}
	\mathcal{M} = \bigcup_{i=1}^c [\frac{1}{K} \sum_{j=1}^K f^\prime(\bm{x^i_j})].
	\label{eq:prototype}
\end{equation}
Besides utilizing the Top-$K$ re-ranking approach to select the prototypes, to further exploit the potential of the test-time scheme, we make the memory iteration learnable, which directly optimizes these prototypes via gradient descent. We construct the learnable memory terms by introducing matrix products of learning representations $f^\prime(\bm{x}_i)$, prototypes $\bm{p}_i$, and pseudo labels $\hat{y}_i$:
\begin{equation}
	\begin{aligned}
		\gamma_i & = f^\prime(\bm{x}_i) \cdot \mathtt{BN}{(\frac{\bm{p}_i}{|\bm{p}_i|} \cdot \hat{y}_i)}, \\
		{{\cal L}_{i}} &= - \frac{1}{N_b}\sum\nolimits_{i = 1}^{N_b}  \gamma_i \log [\frac{\exp (\gamma_i)}{\sum \exp (\gamma_i)}]. \\
	\end{aligned}	
\end{equation}
Therefore, our algorithm can be summarized as follows:
\begin{algorithm}[ht]
	\caption{\Model Algorithm}
	\begin{algorithmic}[1]
		\REQUIRE number of steps $T$; feature extractor $f$ and classifier $q$ pretrained from source domains; the unlabeled target domain.
		\STATE $s = 0$
		\STATE Copy a new model with the source one: $f^{\prime} \leftarrow f, q^{\prime} \leftarrow q$ and freeze $f,q$.
		\FOR{$s \leq  T$}
		\STATE $s = s + 1$
		\STATE Arrived target data $\bm{x}_t$
		\STATE Update Prototypes $\bm{p}$ in ~Eq.~\ref{eq:prototype} using $\bm{x}_t$
		\STATE Calculate the loss $\mathcal{L}_{\text{m}}$ using Eq.~\ref{eq:margin}
		\STATE Calculate the loss $\mathcal{L}_{\text{e}} $ according to Eq.~\ref{eq:ent}.
		\STATE Calculate the overall loss: $\mathcal{L} \gets \mathcal{L}_{\text{i}}+\mathcal{L}_{\text{e}}+\mathcal{L}_{\text{m}}$
		\STATE Update the network models $f^{\prime},q^{\prime}$ using $\mathcal{L}$ in step 10.
		\ENDFOR
	\end{algorithmic}
	\label{alg}
\end{algorithm}

\section{Visualization Results}~\label{sec:vis}
In Figure~\ref{fig:vis-tsne}, we visualize feature embeddings of ERM~\cite{vapnik1998statistical}, TENT~\cite{wang2020tent}, TAST~\cite{jang2022test}, and proposed \Model. Referring to the figure, we find that, feature embeddings extracted by \Model have better intra-class compactness and inter-class separability.
\section{Detailed Results}~\label{sec:supp:exp}
In this section, we provide more detailed experimental results. These results are organized into three parts: 
\begin{itemize}
        \item We provide experiments of transferring BN-based methods to adapt LN layers methods on the VLCS dataset in Table~\ref{tab:supp-ln}.
	\item Table~\ref{tab:supp-1}-Table~\ref{tab:supp-4} demonstrate the comparison between existing test-time methods with ResNet-18~\cite{he2016deep} backbones and \Model on the VLCS~\cite{li2017deeper}, PACS~\cite{torralba2011unbiased}, Office Home~\cite{venkateswara2017deep}, and TerraIncognita~\cite{beery2018recognition} datasets.
	\item  Table~\ref{tab:supp-5}-Table~\ref{tab:supp-8} show the comparison between \Model and the other test-time methods with the ResNet-50~\cite{he2016deep} visual backbones.
	\item  Table~\ref{tab:supp-9}-Table~\ref{tab:supp-12} detail improvements on each domain between base ERM~\cite{vapnik1998statistical} models with different 12 visual backbones on the aforementioned 4 datasets.
\end{itemize}

\begin{table}[ht]
    \centering
    \caption{Comparsion between transferring existing BN-based method and our proposed UniDG with ViT-B16 backbone on the VLCS dataset.}
    \resizebox{0.8\linewidth}{!}{
    \begin{tabular}{lccccc}
        \toprule
        Method & C & L & S & V & Avg  \\ \midrule
         Source & 98.50 & 63.44 & 74.79 & 77.79  & 78.63\\ 
         TENT~\citep{wang2020tent} & 99.23 & 64.36 & 75.23 & 77.94 & 79.19 \\ \hline
         UniDG~(LN) & 99.73 & 66.65 & 78.44 & 79.79 & \reshl{81.15}{2.52} \\ 
         UniDG & 99.91  & 69.60 &   82.86 &     82.82 & \reshl{83.80}{4.61} \\ \hline
    \end{tabular}
    }
    \label{tab:supp-ln}
\end{table}
\begin{table*}[htb]
	\centering 
	\caption{Full results using classifiers trained by ERM for Table 2 on VLCS. We use ResNet-18 as a backbone network.}
	\resizebox{1.0\linewidth}{!}{
		\begin{tabular}{lccccc}
			\toprule
			Method & C & L & S & V & Avg  \\ \midrule
			ERM &  94.70$\pm$1.33 & 63.79$\pm$1.30 & 67.90$\pm$1.97 & 73.15$\pm$1.37 & 74.88 \\ 
			+Tent & 89.82$\pm$2.89 & 61.98$\pm$1.10 & 65.51$\pm$1.91 & 74.21$\pm$1.61 & 72.88  \\ 
			+TentBN & 79.80$\pm$4.74 & 58.51$\pm$1.44 & 61.62$\pm$0.92 & 68.14$\pm$1.74 & 67.02  \\ 
			+TentClf & 94.75$\pm$1.43 & 63.74$\pm$1.41 & 67.92$\pm$2.22 & 65.40$\pm$6.91 & 72.96  \\ 
			+SHOT & 91.45$\pm$6.83 & 48.26$\pm$1.77 & 54.75$\pm$2.59 & 66.51$\pm$1.25 & 65.24  \\
			+SHOTIM & 90.28$\pm$7.00 & 47.96$\pm$1.45 & 54.66$\pm$2.47 & 66.52$\pm$1.19 & 64.86  \\ 
			+PL & 93.57$\pm$2.24 & 53.82$\pm$2.51 & 50.58$\pm$9.50 & 53.91$\pm$2.78 & 62.97  \\ 
			+PLClf & 94.67$\pm$1.38 & 63.64$\pm$1.31 & 67.90$\pm$2.21 & 73.34$\pm$1.00 & 74.89  \\ 
			+T3A & 97.52$\pm$1.99 & 65.32$\pm$2.24 & 70.70$\pm$3.48 & 75.51$\pm$1.75 & 77.26  \\ 
			+TAST  & 99.17$\pm$0.60 & 65.87$\pm$1.90 & 68.13$\pm$1.76 & 75.92$\pm$1.75 & 77.27  \\ 
			+TAST-BN & 92.60$\pm$8.66 & 64.75$\pm$1.29 & 67.27$\pm$3.14 & 76.23$\pm$3.73 & 75.21  \\ \hline
			Base                  & 95.76 $\pm$ 0.00     & 66.31 $\pm$ 0.00     & 70.07 $\pm$ 0.00     & 74.64 $\pm$ 0.00     & 76.7                 \\
			+\Model                        & \reshl{99.79 $\pm$ 0.09}{4.03}      & \reshl{68.78 $\pm$ 0.24}{2.47}      & \reshl{75.01 $\pm$ 0.30}{4.94}      & \reshl{79.93 $\pm$ 0.24}{5.29}      & \reshl{80.88 $\pm$ 0.14}{4.18}  \\
			\bottomrule
		\end{tabular}
	}
	\label{tab:supp-1}
\end{table*}

\begin{table*}[htb]
	\centering
	\caption{Full results using classifiers trained by ERM for Table 2 on PACS. We use ResNet-18 as a backbone network.}
	\resizebox{1.0\linewidth}{!}{
		\begin{tabular}{lccccc} 
			\toprule
			Method & A & C & P & S & Avg  \\ \midrule
			ERM & 77.78$\pm$0.81 & 75.09$\pm$1.22 & 95.19$\pm$0.29 & 69.11$\pm$1.22 & 79.29  \\ 
			+Tent &  82.21$\pm$1.07 & 81.20$\pm$0.51 & 95.32$\pm$0.33 & 76.82$\pm$1.97 & 83.89 \\ 
			+TentBN & 78.89$\pm$0.67 & 77.45$\pm$0.82 & 95.77$\pm$0.40 & 70.89$\pm$2.75 & 80.75 \\ 
			+TentClf & 78.16$\pm$1.05 & 75.01$\pm$1.53 & 95.50$\pm$0.35 & 65.60$\pm$5.96 & 78.57  \\ 
			+SHOT & 81.09$\pm$0.86 & 79.68$\pm$0.91 & 96.18$\pm$0.27 & 72.48$\pm$2.04 & 82.36 \\
			+SHOTIM &  81.10$\pm$0.90 & 79.66$\pm$0.95 & 96.18$\pm$0.27 & 72.35$\pm$2.03 & 82.33   \\ 
			+PL & 76.42$\pm$4.89 & 61.05$\pm$5.48 & 95.70$\pm$0.56 & 50.75$\pm$8.79 & 70.98  \\ 
			+PLClf & 79.09$\pm$1.41 & 75.46$\pm$2.93 & 95.43$\pm$0.32 & 62.48$\pm$7.31 & 78.11  \\ 
			+T3A &   78.81$\pm$0.97 & 77.14$\pm$1.20 & 95.92$\pm$0.36 & 71.44$\pm$1.63 & 80.83 \\ 
			+TAST  &  80.56$\pm$0.53 & 78.26$\pm$0.99 & 96.44$\pm$0.20 & 72.52$\pm$0.77 & 81.94\\ 
			+TAST-BN & 86.49$\pm$0.20 & 83.70$\pm$2.57 & 97.23$\pm$0.11 & 80.85$\pm$1.42 & 87.07  \\ \hline
			Base                  & 76.51 $\pm$ 0.00     & 73.72 $\pm$ 0.00     & 92.37 $\pm$ 0.00     & 76.08 $\pm$ 0.00     & 79.7                 \\
			+\Model                      & \reshl{80.19 $\pm$ 0.13}{3.68}      & \reshl{76.47 $\pm$ 0.37}{2.75}      & \reshl{95.41 $\pm$ 0.15}{3.04}      & \reshl{74.86 $\pm$ 0.38}{-1.22}     & \reshl{81.73 $\pm$ 0.06}{2.03}     
			\\\bottomrule
		\end{tabular}
	}
\end{table*}

\begin{table*}[htb]
	\centering
	\caption{Full results using classifiers trained by ERM for Table 2 on OfficeHome. We use ResNet-18 as a backbone network.}
	\resizebox{1.0\linewidth}{!}{
		\begin{tabular}{lccccc} 
			\toprule
			Method & A & C & P & R & Avg  \\ \midrule
			ERM & 55.19$\pm$0.49 & 47.76$\pm$1.02 & 72.22$\pm$0.53 & 73.21$\pm$0.89 & 62.10  \\ 
			+Tent & 53.39$\pm$0.61 & 48.28$\pm$0.88 & 70.50$\pm$0.68 & 71.29$\pm$0.72 & 60.86 \\ 
			+TentBN & 55.53$\pm$0.43 & 49.53$\pm$0.95 & 72.47$\pm$0.27 & 73.01$\pm$1.23 & 62.64\\ 
			+TentClf & 55.17$\pm$0.67 & 36.73$\pm$1.94 & 72.21$\pm$0.52 & 73.22$\pm$0.97 & 59.33 \\ 
			+SHOT & 55.14$\pm$0.57 & 50.27$\pm$1.18 & 71.69$\pm$0.45 & 73.21$\pm$0.91 & 62.58 \\
			+SHOTIM &  55.08$\pm$0.56 & 50.29$\pm$1.17 & 71.71$\pm$0.40 & 73.21$\pm$0.90 & 62.57  \\ 
			+PL & 54.49$\pm$1.06 & 34.66$\pm$13.13 & 71.45$\pm$0.37 & 72.20$\pm$0.65 & 58.20 \\ 
			+PLClf & 55.14$\pm$0.70 & 47.70$\pm$1.25 & 72.21$\pm$0.54 & 72.62$\pm$0.96 & 61.92  \\ 
			+T3A &  55.10$\pm$0.74 & 49.56$\pm$1.14 & 74.10$\pm$0.55 & 74.07$\pm$1.18 & 63.21 \\ \midrule
			+TAST  & 56.15$\pm$0.68 & 50.04$\pm$1.31 & 74.33$\pm$0.28 & 74.28$\pm$1.23 & 63.70  \\ 
			+TAST-BN & 55.11$\pm$0.58 & 51.35$\pm$0.85 & 72.58$\pm$0.80 & 72.13$\pm$0.78 & 62.79  \\ 
			\hline
			Base                  & 46.91 $\pm$ 0.00     & 45.42 $\pm$ 0.00     & 65.06 $\pm$ 0.00     & 66.49 $\pm$ 0.00     & 56.0                 \\
			+\Model                      & \reshl{49.02 $\pm$ 0.11}{2.11}      & \reshl{47.33 $\pm$ 0.24}{1.91}      & \reshl{69.14 $\pm$ 0.15}{4.08}      & \reshl{68.17 $\pm$ 0.22}{1.68}      & \reshl{58.42 $\pm$ 0.06}{2.42}  
			\\ \bottomrule
		\end{tabular}
	}
\end{table*}

\begin{table*}[htb]
	\centering
	\caption{Full results using classifiers trained by ERM for Table 2 on TerraIncognita. We use ResNet-18 as a backbone network.}
	\resizebox{1.0\linewidth}{!}{
		\begin{tabular}{lccccc} 
			\toprule
			Method & L100 & L38 & L43 & L46 & Avg  \\ \midrule
			ERM & 37.18$\pm$2.46 & 36.12$\pm$4.20 & 53.18$\pm$1.27 & 36.02$\pm$1.37 & 40.62   \\ 
			+Tent & 38.29$\pm$0.48 & 25.82$\pm$3.91 & 41.53$\pm$1.59 & 29.15$\pm$1.83 & 33.70 \\ 
			+TentBN & 40.55$\pm$1.46 & 37.44$\pm$2.22 & 46.33$\pm$1.32 & 35.30$\pm$1.26 & 39.91 \\ 
			+TentClf & 34.44$\pm$13.31 & 34.19$\pm$5.76 & 52.71$\pm$2.03 & 31.86$\pm$2.26 & 38.30 \\ 
			+SHOT & 33.87$\pm$0.66 & 28.58$\pm$2.10 & 40.99$\pm$2.07 & 30.83$\pm$1.26 & 33.57 \\
			+SHOTIM &  33.83$\pm$1.29 & 28.13$\pm$2.30 & 40.81$\pm$2.18 & 30.64$\pm$1.46 & 33.35 \\ 
			+PL & 51.92$\pm$1.19 & 35.61$\pm$20.74 & 39.97$\pm$10.98 & 22.26$\pm$8.21 & 37.44\\ 
			+PLClf & 45.22$\pm$2.45 & 36.03$\pm$5.81 & 52.76$\pm$1.54 & 33.10$\pm$2.27 & 41.78 \\ 
			+T3A & 36.22$\pm$1.89 & 40.08$\pm$1.98 & 50.72$\pm$1.02 & 33.79$\pm$1.25 & 40.20   \\ 
			+TAST  & 43.67$\pm$2.83 & 39.24$\pm$3.79 & 52.64$\pm$3.02 & 35.01$\pm$1.09 & 42.64 \\ 
			+TAST-BN & 51.06$\pm$7.31 & 32.74$\pm$7.54 & 41.70$\pm$2.86 & 32.21$\pm$3.05 & 39.43 \\ \midrule
			Base                  & 45.03 $\pm$ 0.00     & 32.67 $\pm$ 0.00     & 47.54 $\pm$ 0.00     & 35.80 $\pm$ 0.00     & 40.3                 \\
			+\Model                      & \reshl{54.34 $\pm$ 0.21}{9.31}      & \reshl{51.48 $\pm$ 2.16}{18.81}     & \reshl{48.27 $\pm$ 0.46}{0.73}      & \reshl{37.64 $\pm$ 0.21}{1.84}      & \reshl{47.93 $\pm$ 0.65}{7.63} 
			\\ \bottomrule
		\end{tabular}
	}
	\label{tab:supp-4}
\end{table*}

\begin{table*}[htb]
	\centering
	\caption{Full results using classifiers trained by ERM for Table 2 on VLCS. We use ResNet-50 as a backbone network.}
	\resizebox{1.0\linewidth}{!}{
		\begin{tabular}{lccccc}
			\toprule
			Method & C & L & S & V & Avg  \\ \midrule
			ERM & 97.66$\pm$0.64 & 63.87$\pm$1.71 & 71.21$\pm$1.52 & 74.09$\pm$2.06  & 76.71  \\ 
			+Tent & 92.36$\pm$2.44 & 58.46$\pm$3.29 & 67.84$\pm$2.03 & 73.19$\pm$2.68  & 72.96  \\ 
			+TentBN & 85.36$\pm$3.49 & 58.35$\pm$3.46 & 66.47$\pm$2.71 & 68.42$\pm$2.11  & 69.65 \\ 
			+TentClf & 97.61$\pm$0.58 & 63.67$\pm$2.10 & 68.77$\pm$1.27 & 73.16$\pm$1.31  & 75.80 \\ 
			+SHOT & 98.72$\pm$1.50 & 46.82$\pm$2.57 & 55.70$\pm$1.78 & 67.04$\pm$2.88  & 67.07 \\
			+SHOTIM & 98.65$\pm$1.46 & 46.54$\pm$2.32 & 55.81$\pm$2.32 & 66.73$\pm$2.82  & 66.93  \\ 
			+PL & 98.48$\pm$0.34 & 53.45$\pm$2.82 & 59.45$\pm$9.24 & 66.24$\pm$8.63  & 69.41  \\ 
			+PLClf &  97.63$\pm$0.64 & 63.36$\pm$2.10 & 69.74$\pm$0.78 & 71.86$\pm$4.53  & 75.65 \\ 
			+T3A & 99.17$\pm$0.38 & 64.78$\pm$1.61 & 73.01$\pm$3.24 & 72.20$\pm$2.84  & 77.29   \\ 
			+TAST  & 99.35$\pm$0.30 & 65.64$\pm$1.78 & 73.63$\pm$3.58 & 72.01$\pm$2.68  & 77.66  \\ 
			+TAST-BN & 96.09$\pm$2.40 & 60.22$\pm$6.08 & 65.78$\pm$6.51 & 71.99$\pm$5.90  & 73.52  \\ \midrule
			Base                  & 97.70 $\pm$ 0.00     & 63.20 $\pm$ 0.00     & 70.18 $\pm$ 0.00     & 78.82 $\pm$ 0.00     & 77.50                 \\
			+\Model                      & \reshl{99.71 $\pm$ 0.05}{2.01}      & \reshl{71.11 $\pm$ 0.02}{7.91}      & \reshl{73.60 $\pm$ 0.38}{3.42}      & \reshl{81.99 $\pm$ 0.06}{3.17}      & \reshl{81.60 $\pm$ 0.08}{4.10}
			\\ 
			\bottomrule
		\end{tabular}
	}
	\label{tab:supp-5}
\end{table*}

\begin{table*}[htb]
	\centering
	\caption{Full results using classifiers trained by ERM for Table 2 on PACS. We use ResNet-50 as a backbone network.}
	\resizebox{1.0\linewidth}{!}{
		\begin{tabular}{lccccc} 
			\toprule
			Method &A&C&P&S& Avg  \\ \midrule
			ERM & 82.92$\pm$1.65 & 78.05$\pm$3.36 & 96.50$\pm$0.32 & 75.38$\pm$3.31 & 83.21  \\ 
			+Tent & 82.54$\pm$1.32 & 84.90$\pm$1.35 & 95.45$\pm$0.93 & 77.74$\pm$1.36 & 85.16   \\ 
			+TentBN & 82.75$\pm$2.01 & 79.50$\pm$2.26 & 96.78$\pm$0.20 & 75.73$\pm$3.22 & 83.69 \\ 
			+TentClf & 83.00$\pm$1.87 & 77.86$\pm$4.20 & 96.55$\pm$0.36 & 73.25$\pm$6.14 & 82.66   \\ 
			+SHOT & 84.67$\pm$1.70 & 80.17$\pm$1.39 & 96.58$\pm$0.52 & 74.86$\pm$2.95 & 84.07 \\
			+SHOTIM &  84.62$\pm$1.79 & 80.24$\pm$1.41 & 96.54$\pm$0.46 & 75.16$\pm$2.88 & 84.14 \\ 
			+PL & 84.59$\pm$5.51 & 76.35$\pm$2.57 & 96.41$\pm$0.68 & 69.54$\pm$11.22 & 81.72  \\ 
			+PLClf & 83.88$\pm$2.00 & 78.93$\pm$3.68 & 96.53$\pm$0.40 & 73.96$\pm$6.08 & 83.33  \\ 
			+T3A & 83.56$\pm$2.03 & 79.75$\pm$3.14 & 96.99$\pm$0.24 & 75.36$\pm$3.57 & 83.92   \\ 
			+TAST  & 83.85$\pm$2.05 & 79.15$\pm$3.03 & 96.93$\pm$0.27 & 76.49$\pm$3.13 & 84.11 \\ 
			+TAST-BN &  87.11$\pm$2.04 & 88.50$\pm$1.93 & 97.79$\pm$0.47 & 83.23$\pm$1.42 & 89.16  \\ \midrule
			Base                  & 86.52 $\pm$ 0.00     & 71.75 $\pm$ 0.00     & 97.16 $\pm$ 0.00     & 75.57 $\pm$ 0.00     & 82.7                 \\
			+\Model                      & \reshl{91.26 $\pm$ 0.11}{4.74}      & \reshl{84.86 $\pm$ 0.63}{13.11}     & \reshl{98.13 $\pm$ 0.05}{0.97}      & \reshl{81.77 $\pm$ 0.56}{6.2}       & \reshl{89.00 $\pm$ 0.30}{6.30} 
			\\ \bottomrule
		\end{tabular}
	}
\end{table*}

\begin{table*}[htb]
	\centering
	\caption{Full results using classifiers trained by ERM for Table 2 on OfficeHome. We use ResNet-50 as a backbone network.}
	\resizebox{1.0\linewidth}{!}{
		\begin{tabular}{lccccc} 
			\toprule
			Method & A&C&P&R & Avg  \\ \midrule
			ERM & 61.32$\pm$0.69 & 53.44$\pm$1.11 & 75.84$\pm$1.10 & 77.90$\pm$0.92 & 67.13  \\ 
			+Tent & 60.98$\pm$0.67 & 53.94$\pm$1.24 & 74.49$\pm$0.71 & 75.75$\pm$0.53 & 66.29  \\ 
			+TentBN & 62.63$\pm$0.45 & 54.90$\pm$1.17 & 76.20$\pm$1.09 & 77.92$\pm$1.01 & 67.91 \\ 
			+TentClf & 61.35$\pm$0.73 & 52.72$\pm$1.40 & 75.23$\pm$1.05 & 77.86$\pm$1.07 & 66.79 \\ 
			+SHOT & 61.91$\pm$0.33 & 55.58$\pm$0.91 & 75.49$\pm$1.54 & 77.60$\pm$0.80 & 67.65  \\
			+SHOTIM & 61.84$\pm$0.32 & 55.63$\pm$0.92 & 75.56$\pm$1.60 & 77.57$\pm$0.79 & 67.65  \\ 
			+PL &  59.42$\pm$1.55 & 42.40$\pm$12.31 & 73.80$\pm$2.26 & 75.77$\pm$1.50 & 62.85   \\ 
			+PLClf & 61.35$\pm$0.40 & 52.87$\pm$1.96 & 75.86$\pm$1.09 & 77.94$\pm$1.10 & 67.01   \\ 
			+T3A &  61.91$\pm$0.59 & 55.07$\pm$1.14 & 77.39$\pm$1.38 & 78.67$\pm$0.61 & 68.26   \\ 
			+TAST  & 62.43$\pm$0.80 & 55.81$\pm$1.26 & 77.46$\pm$1.07 & 78.83$\pm$0.93 & 68.63  \\ 
			+TAST-BN &  63.22$\pm$0.85 & 58.20$\pm$0.98 & 77.14$\pm$1.10 & 76.94$\pm$0.39 & 68.88 \\ \midrule
			Base                  & 57.67 $\pm$ 0.00     & 53.69 $\pm$ 0.00     & 73.90 $\pm$ 0.00     & 75.70 $\pm$ 0.00     & 65.2                 \\
			+\Model                      & \reshl{62.84 $\pm$ 0.31}{5.17}      & \reshl{55.86 $\pm$ 0.12}{2.17}      & \reshl{78.16 $\pm$ 0.05}{4.26}      & \reshl{78.69 $\pm$ 0.02}{2.99}      & \reshl{68.89 $\pm$ 0.07}{3.69}
			\\ \bottomrule 
		\end{tabular}
	}
\end{table*}

\begin{table*}[htb]
	\centering
	\caption{Full results using classifiers trained by ERM for Table 2 on TerraIncognita. We use ResNet-50 as a backbone network.}
	\resizebox{1.0\linewidth}{!}{
		\begin{tabular}{lccccc} 
			\toprule
			Method & L100&L38&L43&L46 & Avg  \\ \midrule
			ERM & 46.84$\pm$1.96 & 43.24$\pm$2.51 & 53.32$\pm$1.92 & 40.30$\pm$1.93 & 45.93  \\ 
			+Tent & 41.20$\pm$2.71 & 29.72$\pm$3.59 & 41.35$\pm$2.92 & 36.03$\pm$2.85 & 37.08  \\ 
			+TentBN & 46.64$\pm$1.17 & 41.11$\pm$3.16 & 49.31$\pm$1.05 & 38.52$\pm$2.04 & 43.89 \\ 
			+TentClf & 49.87$\pm$3.80 & 43.31$\pm$3.19 & 53.01$\pm$2.31 & 28.40$\pm$6.19 & 43.64 \\ 
			+SHOT & 36.17$\pm$2.70 & 29.80$\pm$2.92 & 41.00$\pm$0.30& 33.83$\pm$1.86 & 35.20  \\
			+SHOTIM & 35.56$\pm$2.76 & 27.49$\pm$4.01 & 40.77$\pm$0.45 & 33.67$\pm$1.84 & 34.37   \\ 
			+PL & 56.75$\pm$5.78 & 46.12$\pm$1.03 & 29.44$\pm$10.14 & 20.06$\pm$4.65 & 38.09 \\ 
			+PLClf &  52.28$\pm$3.95 & 43.76$\pm$2.96 & 52.78$\pm$2.15 & 37.81$\pm$2.49 & 46.66   \\ 
			+T3A & 45.13$\pm$1.26 & 44.67$\pm$2.56 & 52.52$\pm$0.78 & 40.13$\pm$2.31 & 45.61    \\ 
			+TAST  & 53.01$\pm$3.95 & 43.27$\pm$3.21 & 53.79$\pm$2.72 & 39.66$\pm$3.65 & 47.43 \\ 
			+TAST-BN & 55.75$\pm$2.37 & 33.92$\pm$9.86 & 43.87$\pm$4.70 & 32.33$\pm$4.40 & 41.47   \\ \midrule
			Base                  & 49.70 $\pm$ 0.00     & 40.51 $\pm$ 0.00     & 56.45 $\pm$ 0.00     & 33.63 $\pm$ 0.00     & 45.1                 \\
			+\Model                      & \reshl{64.37 $\pm$ 0.11}{14.67}     & \reshl{46.87 $\pm$ 0.32}{6.36}      & \reshl{57.37 $\pm$ 0.22}{0.92}      & \reshl{42.82 $\pm$ 0.57}{9.19}      & \reshl{52.86 $\pm$ 0.17}{7.76} 
			\\ \bottomrule
		\end{tabular}
	}
	\label{tab:supp-8}
\end{table*}
\begin{table*}[htb]
	\centering 
	\caption{Full results using classifiers trained by ERM with different visual backbones for Table 3 on the VLCS dataset. }
	\resizebox{1.0\linewidth}{!}{
		\begin{tabular}{lccccc}
			\toprule
			Method & C & L & S & V & Avg  \\ \midrule
			ResNet-101~\cite{he2016deep}                  & 97.00 $\pm$ 0.00     & 65.22 $\pm$ 0.00     & 70.64 $\pm$ 0.00     & 73.60 $\pm$ 0.00     & 76.6                 \\
			+\Model                       & \reshl{99.35 $\pm$ 0.05}{2.35}      & \reshl{69.71 $\pm$ 0.30}{4.49}      & \reshl{75.01 $\pm$ 0.13}{4.37}      & \reshl{78.05 $\pm$ 0.30}{4.45}      & \reshl{80.53 $\pm$ 0.16}{3.93}         \\ \midrule			 
			ViT-B16~\cite{dosovitskiy2020image}                  & 98.50 $\pm$ 0.00     & 63.44 $\pm$ 0.00     & 74.79 $\pm$ 0.00     & 77.67 $\pm$ 0.00     & 78.6                 \\
			+\Model                      & \reshl{99.94 $\pm$ 0.02}{1.44}      & \reshl{70.07 $\pm$ 0.17}{6.63}      & \reshl{82.20 $\pm$ 0.63}{7.41}      & \reshl{82.07 $\pm$ 0.10}{4.4}       & \reshl{83.57 $\pm$ 0.13}{4.97} \\
			\midrule
			DeiT~\cite{touvron2021training}                   & 97.08 $\pm$ 0.00     & 66.96 $\pm$ 0.00     & 74.64 $\pm$ 0.00     & 79.27 $\pm$ 0.00     & 79.5                 \\
			+\Model                      & \reshl{99.94 $\pm$ 0.02}{2.86}      & \reshl{69.49 $\pm$ 0.19}{2.53}      & \reshl{83.69 $\pm$ 0.13}{9.05}      & \reshl{87.39 $\pm$ 0.28}{8.12}      & \reshl{85.13 $\pm$ 0.05}{5.63}      \\
			\midrule
			HViT~\cite{dosovitskiy2020image}                  & 96.73 $\pm$ 0.00     & 64.38 $\pm$ 0.00     & 75.40 $\pm$ 0.00     & 79.90 $\pm$ 0.00     & 79.1                 \\
			+\Model                      & \reshl{99.53 $\pm$ 0.09}{2.8}       & \reshl{69.49 $\pm$ 0.05}{5.11}      & \reshl{79.27 $\pm$ 0.16}{3.87}      & \reshl{85.82 $\pm$ 0.22}{5.92}      & \reshl{83.53 $\pm$ 0.12}{4.43}      \\
			\midrule
			Swin Transformer~\cite{liu2021swin}                   & 94.79 $\pm$ 0.00     & 65.27 $\pm$ 0.00     & 79.82 $\pm$ 0.00     & 80.04 $\pm$ 0.00     & 80.0                 \\
			+\Model                      & \reshl{100.00 $\pm$ 0.00}{5.21}     & \reshl{69.51 $\pm$ 0.35}{4.24}      & \reshl{83.79 $\pm$ 0.32}{3.97}      & \reshl{86.81 $\pm$ 0.15}{6.77}      & \reshl{85.03 $\pm$ 0.10}{5.03}      \\
			\midrule
			MobileNet V3~\cite{howard2019searching}                  & 72.8 $\pm$ 0.0       & 57.6 $\pm$ 0.0       & 58.3 $\pm$ 0.0  & 74.4  $\pm$ 0.0         & 65.8                 \\
			+\Model                      & \reshl{93.8 $\pm$ 0.3}{21.0}        & \reshl{63.2 $\pm$ 0.1}{5.6}         & \reshl{69.1 $\pm$ 0.2}{10.8}        & \reshl{78.8 $\pm$ 0.1}{4.4}         & \reshl{76.2 $\pm$ 0.1}{10.4}         \\
			\midrule
			ConvNeXt~\cite{liu2022convnet}                  & 97.97 $\pm$ 0.00     & 68.09 $\pm$ 0.00     & 78.37 $\pm$ 0.00     & 73.64 $\pm$ 0.00     & 79.5                 \\
			+\Model                      & \reshl{100.00 $\pm$ 0.00}{2.03}     & \reshl{72.97 $\pm$ 0.69}{4.88}      & \reshl{84.77 $\pm$ 0.09}{6.4}       & \reshl{85.50 $\pm$ 0.47}{11.86}     & \reshl{85.81 $\pm$ 0.25}{6.31}      \\
			\midrule
			ViT-L16~\cite{dosovitskiy2020image}                  & 97.88 $\pm$ 0.00     & 61.74 $\pm$ 0.00     & 71.90 $\pm$ 0.00     & 73.94 $\pm$ 0.00     & 76.4                 \\
			+\Model                      & \reshl{100.00 $\pm$ 0.00}{2.12}     & \reshl{69.05 $\pm$ 0.38}{7.31}      & \reshl{77.89 $\pm$ 0.55}{5.99}      & \reshl{85.71 $\pm$ 0.44}{11.77}     & \reshl{83.16 $\pm$ 0.23}{6.76}      \\
			\midrule
			Mixer-B16~\cite{tolstikhin2021mlp}                  & 93.46 $\pm$ 0.00     & 58.87 $\pm$ 0.00     & 70.26 $\pm$ 0.00     & 73.23 $\pm$ 0.00     & 74.0                 \\
			+\Model                      & \reshl{99.20 $\pm$ 0.08}{5.74}      & \reshl{67.69 $\pm$ 0.47}{8.82}      & \reshl{79.14 $\pm$ 0.74}{8.88}      & \reshl{79.13 $\pm$ 0.18}{5.9}       & \reshl{81.29 $\pm$ 0.23}{7.29}      \\
			\midrule
			Mixer-L16~\cite{tolstikhin2021mlp}                  & 97.61 $\pm$ 0.00     & 61.65 $\pm$ 0.00     & 75.55 $\pm$ 0.00     & 77.42 $\pm$ 0.00     & 78.1                 \\
			+\Model                      & \reshl{99.91 $\pm$ 0.04}{2.3}       & \reshl{65.80 $\pm$ 0.28}{4.15}      & \reshl{82.53 $\pm$ 0.06}{6.98}      & \reshl{83.94 $\pm$ 0.05}{6.52}      & \reshl{83.05 $\pm$ 0.09}{4.95}      \\
			\bottomrule
		\end{tabular}
	}
	\label{tab:supp-9}
\end{table*}

\begin{table*}[htb]
	\centering
	\caption{Full results using classifiers trained by ERM with different visual backbones for Table 3 on PACS dataset.}
	\resizebox{1.0\linewidth}{!}{
		\begin{tabular}{lccccc} 
			\toprule
			Method & A & C & P & S & Avg  \\ \midrule
			ResNet-101~\cite{he2016deep}                  & 83.83 $\pm$ 0.00     & 82.30 $\pm$ 0.00     & 97.98 $\pm$ 0.00     & 79.29 $\pm$ 0.00     & 85.9                 \\
			+\Model                        & \reshl{87.29 $\pm$ 0.14}{3.46}      & \reshl{85.61 $\pm$ 0.09}{3.31}      & \reshl{98.58 $\pm$ 0.00}{0.6}       & \reshl{81.91 $\pm$ 0.43}{2.62}      & \reshl{88.35 $\pm$ 0.06}{2.45}      \\
			\midrule	
			ViT-B16~\cite{dosovitskiy2020image}                  & 85.54 $\pm$ 0.00     & 82.62 $\pm$ 0.00     & 98.88 $\pm$ 0.00     & 54.20 $\pm$ 0.00     & 80.3                 \\
			+\Model                      & \reshl{89.85 $\pm$ 0.20}{4.31}      & \reshl{87.28 $\pm$ 0.15}{4.66}      & \reshl{99.78 $\pm$ 0.04}{0.9}       & \reshl{64.84 $\pm$ 2.01}{10.64}     & \reshl{85.44 $\pm$ 0.54}{5.14}      \\
			\midrule	
			DeiT~\cite{touvron2021training}                   & 90.73 $\pm$ 0.00     & 83.74 $\pm$ 0.00     & 99.03 $\pm$ 0.00     & 77.99 $\pm$ 0.00     & 87.9                 \\
			+\Model                      & \reshl{94.55 $\pm$ 0.18}{3.82}      & \reshl{90.60 $\pm$ 0.51}{6.86}      & \reshl{99.70 $\pm$ 0.00}{0.67}      & \reshl{85.41 $\pm$ 0.55}{7.42}      & \reshl{92.57 $\pm$ 0.31}{4.67}      \\
			\midrule
			HViT~\cite{dosovitskiy2020image}                  & 93.53 $\pm$ 0.00     & 81.66 $\pm$ 0.00     & 98.65 $\pm$ 0.00     & 85.56 $\pm$ 0.00     & 89.9                 \\
			+\Model                      & \reshl{96.24 $\pm$ 0.12}{2.71}      & \reshl{88.84 $\pm$ 0.33}{7.18}      & \reshl{99.63 $\pm$ 0.04}{0.98}      & \reshl{89.26 $\pm$ 0.08}{3.7}       & \reshl{93.49 $\pm$ 0.10}{3.59}      \\
			\midrule
			Swin Transformer~\cite{liu2021swin}                   & 93.11 $\pm$ 0.00     & 85.82 $\pm$ 0.00     & 99.48 $\pm$ 0.00     & 83.97 $\pm$ 0.00     & 90.6                 \\
			+\Model                      & \reshl{97.38 $\pm$ 0.03}{4.27}      & \reshl{90.65 $\pm$ 0.49}{4.83}      & \reshl{99.78 $\pm$ 0.04}{0.3}       & \reshl{89.50 $\pm$ 0.30}{5.53}      & \reshl{94.33 $\pm$ 0.16}{3.73}      \\
			\midrule
			MobileNet V3~\cite{howard2019searching}                  & 78.8 $\pm$ 0.0       & 77.7 $\pm$ 0.0       & 94.2 $\pm$ 0.0       & 66.5 $\pm$ 0.0       & 79.3                 \\
			+\Model                      & \reshl{84.9 $\pm$ 0.2}{6.1}        & \reshl{83.5 $\pm$ 0.2}{5.8}        & \reshl{97.7 $\pm$ 0.1}{3.5}         & \reshl{74.8 $\pm$ 0.3}{8.3}         & \reshl{85.3 $\pm$ 0.4}{6.0}         \\
			\midrule
			ConvNeXt~\cite{liu2022convnet}                  & 95.73 $\pm$ 0.00     & 85.34 $\pm$ 0.00     & 99.25 $\pm$ 0.00     & 90.81 $\pm$ 0.00     & 92.8                 \\
			+\Model                      & \reshl{98.21 $\pm$ 0.09}{2.48}      & \reshl{90.23 $\pm$ 0.50}{4.89}      & \reshl{99.93 $\pm$ 0.00}{0.68}      & \reshl{92.88 $\pm$ 0.16}{2.07}      & \reshl{95.31 $\pm$ 0.16}{2.51}      \\
			\midrule
			ViT-L16~\cite{dosovitskiy2020image}                  & 93.17 $\pm$ 0.00     & 88.06 $\pm$ 0.00     & 99.55 $\pm$ 0.00     & 83.30 $\pm$ 0.00     & 91.0                 \\
			+\Model                      & \reshl{98.19 $\pm$ 0.02}{5.02}      & \reshl{92.54 $\pm$ 0.11}{4.48}      & \reshl{99.98 $\pm$ 0.02}{0.43}      & \reshl{90.21 $\pm$ 0.36}{6.91}      & \reshl{95.23 $\pm$ 0.12}{4.23}      \\
			\midrule
			Mixer-B16~\cite{tolstikhin2021mlp}                  & 76.02 $\pm$ 0.00     & 71.59 $\pm$ 0.00     & 94.39 $\pm$ 0.00     & 62.09 $\pm$ 0.00     & 76.0                 \\
			+\Model                      & \reshl{84.16 $\pm$ 0.35}{8.14}      & \reshl{82.16 $\pm$ 0.38}{10.57}     & \reshl{96.96 $\pm$ 0.05}{2.57}      & \reshl{65.91 $\pm$ 0.10}{3.82}      & \reshl{82.30 $\pm$ 0.13}{6.3}       \\
			\midrule
			Mixer-L16~\cite{tolstikhin2021mlp}                  & 84.93 $\pm$ 0.00     & 82.89 $\pm$ 0.00     & 98.73 $\pm$ 0.00     & 73.66 $\pm$ 0.00     & 85.1                 \\
			+\Model                      & \reshl{91.74 $\pm$ 0.16}{6.81}      & \reshl{85.89 $\pm$ 0.38}{3.0}       & \reshl{99.38 $\pm$ 0.07}{0.65}      & \reshl{76.86 $\pm$ 0.33}{3.2}       & \reshl{88.47 $\pm$ 0.16}{3.37}      
			\\\bottomrule
		\end{tabular}
	}
\end{table*}

\begin{table*}[htb]
	\centering
	\caption{Full results using classifiers trained by ERM with different visual backbones for Table 3 on OfficeHome dataset.}
	\resizebox{1.0\linewidth}{!}{
		\begin{tabular}{lccccc} 
			\toprule
			Method & A & C & P & R & Avg  \\ \midrule
			ResNet-101~\cite{he2016deep}                  & 61.23 $\pm$ 0.00     & 55.18 $\pm$ 0.00     & 75.56 $\pm$ 0.00     & 77.74 $\pm$ 0.00     & 67.4                 \\
			+\Model                      & \reshl{65.00 $\pm$ 0.28}{3.77}      & \reshl{58.17 $\pm$ 0.51}{2.99}      & \reshl{78.60 $\pm$ 0.12}{3.04}      & \reshl{79.54 $\pm$ 0.02}{1.8}       & \reshl{70.33 $\pm$ 0.20}{2.93}      \\
			\midrule	
			ViT-B16~\cite{dosovitskiy2020image}                  & 71.37 $\pm$ 0.00     & 60.17 $\pm$ 0.00     & 84.40 $\pm$ 0.00     & 86.60 $\pm$ 0.00     & 75.6                 \\
			+\Model                      & \reshl{78.29 $\pm$ 0.32}{6.92}      & \reshl{66.46 $\pm$ 0.39}{6.29}      & \reshl{89.10 $\pm$ 0.16}{4.7}       & \reshl{89.99 $\pm$ 0.12}{3.39}      & \reshl{80.96 $\pm$ 0.05}{5.36}      \\
			\midrule
			DeiT~\cite{touvron2021training}                   & 75.28 $\pm$ 0.00     & 62.69 $\pm$ 0.00     & 84.23 $\pm$ 0.00     & 85.86 $\pm$ 0.00     & 77.0                 \\
			+\Model                      & \reshl{78.30 $\pm$ 0.07}{3.02}      & \reshl{66.56 $\pm$ 0.46}{3.87}      & \reshl{85.82 $\pm$ 0.07}{1.59}      & \reshl{87.14 $\pm$ 0.06}{1.28}      & \reshl{79.46 $\pm$ 0.10}{2.46}      \\
			\midrule
			HViT~\cite{dosovitskiy2020image}                  & 74.67 $\pm$ 0.00     & 70.85 $\pm$ 0.00     & 86.37 $\pm$ 0.00     & 87.44 $\pm$ 0.00     & 79.8                 \\
			+\Model                      & \reshl{76.67 $\pm$ 0.11}{2.0}       & \reshl{72.27 $\pm$ 0.54}{1.42}      & \reshl{87.33 $\pm$ 0.14}{0.96}      & \reshl{89.08 $\pm$ 0.22}{1.64}      & \reshl{81.34 $\pm$ 0.06}{1.54}      \\
			\midrule
			Swin Transformer~\cite{liu2021swin}                   & 80.74 $\pm$ 0.00     & 72.22 $\pm$ 0.00     & 88.20 $\pm$ 0.00     & 89.30 $\pm$ 0.00     & 82.6                 \\
			+\Model                      & \reshl{83.26 $\pm$ 0.17}{2.52}      & \reshl{74.42 $\pm$ 0.03}{2.2}       & \reshl{89.90 $\pm$ 0.10}{1.7}       & \reshl{90.69 $\pm$ 0.05}{1.39}      & \reshl{84.57 $\pm$ 0.06}{1.97}      \\
			\midrule
			MobileNet V3~\cite{howard2019searching}                  & 56.7 $\pm$ 0.0       & 46.2 $\pm$ 0.0       & 68.9 $\pm$ 0.0       & 73.0 $\pm$ 0.0       & 61.2                 \\
			+\Model                      & \reshl{61.2 $\pm$ 0.5}{4.5}         & \reshl{50.7 $\pm$ 0.5}{4.5}        & \reshl{73.1 $\pm$ 0.1}{4.2}        & \reshl{75.4 $\pm$ 0.1}{2.4}        & \reshl{65.1 $\pm$ 0.2}{3.9}        \\
			\midrule
			ConvNeXt~\cite{liu2022convnet}                  & 83.88 $\pm$ 0.00     & 78.12 $\pm$ 0.00     & 90.62 $\pm$ 0.00     & 91.37 $\pm$ 0.00     & 86.0                 \\
			+\Model                      & \reshl{87.68 $\pm$ 0.11}{3.8}       & \reshl{80.95 $\pm$ 0.07}{2.83}      & \reshl{92.25 $\pm$ 0.04}{1.63}      & \reshl{93.23 $\pm$ 0.09}{1.86}      & \reshl{88.53 $\pm$ 0.04}{2.53}      \\
			\midrule
			ViT-L16~\cite{dosovitskiy2020image}                  & 81.31 $\pm$ 0.00     & 71.56 $\pm$ 0.00     & 89.33 $\pm$ 0.00     & 90.71 $\pm$ 0.00     & 83.2                 \\
			+\Model                      & \reshl{86.25 $\pm$ 0.29}{4.94}      & \reshl{79.71 $\pm$ 0.20}{8.15}      & \reshl{91.35 $\pm$ 0.16}{2.02}      & \reshl{92.85 $\pm$ 0.11}{2.14}      & \reshl{87.54 $\pm$ 0.17}{4.34}      \\
			\midrule
			Mixer-B16~\cite{tolstikhin2021mlp}                  & 30.69 $\pm$ 0.00     & 47.97 $\pm$ 0.00     & 70.33 $\pm$ 0.00     & 60.90 $\pm$ 0.00     & 52.5                 \\
			+\Model                      & \reshl{32.60 $\pm$ 0.24}{1.91}      & \reshl{53.68 $\pm$ 0.73}{5.71}      & \reshl{77.18 $\pm$ 0.13}{6.85}      & \reshl{67.26 $\pm$ 0.22}{6.36}      & \reshl{57.68 $\pm$ 0.27}{5.18}      \\
			\midrule
			Mixer-L16~\cite{tolstikhin2021mlp}                  & 67.61 $\pm$ 0.00     & 62.54 $\pm$ 0.00     & 82.49 $\pm$ 0.00     & 68.10 $\pm$ 0.00     & 70.2                 \\
			+\Model                      & \reshl{71.78 $\pm$ 0.36}{4.17}      & \reshl{68.41 $\pm$ 0.23}{5.87}      & \reshl{87.08 $\pm$ 0.15}{4.59}      & \reshl{75.04 $\pm$ 0.16}{6.94}      & \reshl{75.58 $\pm$ 0.08}{5.38}     
			\\ \bottomrule
		\end{tabular}
	}
\end{table*}

\begin{table*}[htb]
	\centering
	\caption{Full results using classifiers trained by ERM with different visual backbones for Table 3 on TerraIncognita dataset.}
	\resizebox{1.0\linewidth}{!}{
		\begin{tabular}{lccccc} 
			\toprule
			Method & L100 & L38 & L43 & L46 & Avg  \\ \midrule
			ResNet-101~\cite{he2016deep}                  & 37.44 $\pm$ 0.00     & 36.78 $\pm$ 0.00     & 57.08 $\pm$ 0.00     & 39.11 $\pm$ 0.00     & 42.6                 \\
			+\Model                      & \reshl{50.65 $\pm$ 0.47}{13.21}     & \reshl{46.71 $\pm$ 0.76}{9.93}      & \reshl{59.87 $\pm$ 0.58}{2.79}      & \reshl{42.89 $\pm$ 0.85}{3.78}      & \reshl{50.03 $\pm$ 0.46}{7.43}      \\
			\midrule	
			ViT-B16~\cite{dosovitskiy2020image}                  & 53.97 $\pm$ 0.00     & 36.12 $\pm$ 0.00     & 48.46 $\pm$ 0.00     & 35.25 $\pm$ 0.00     & 43.4                 \\
			+\Model                      & \reshl{65.50 $\pm$ 1.36}{11.53}     & \reshl{48.83 $\pm$ 1.34}{12.71}     & \reshl{51.61 $\pm$ 0.27}{3.15}      & \reshl{39.68 $\pm$ 0.16}{4.43}      & \reshl{51.40 $\pm$ 0.22}{8.0}\\
			\midrule
			DeiT~\cite{touvron2021training}                   & 56.45 $\pm$ 0.00     & 42.20 $\pm$ 0.00     & 55.07 $\pm$ 0.00     & 43.40 $\pm$ 0.00     & 49.3                 \\
			+\Model                      & \reshl{64.56 $\pm$ 0.60}{8.11}      & \reshl{51.00 $\pm$ 1.93}{8.8}       & \reshl{56.55 $\pm$ 0.52}{1.48}      & \reshl{44.32 $\pm$ 0.13}{0.92}      & \reshl{54.11 $\pm$ 0.42}{4.81}\\
			\midrule
			HViT~\cite{dosovitskiy2020image}                  & 61.59 $\pm$ 0.00     & 46.13 $\pm$ 0.00     & 61.56 $\pm$ 0.00     & 42.11 $\pm$ 0.00     & 52.8                 \\
			+\Model                      & \reshl{73.64 $\pm$ 1.23}{12.05}     & \reshl{54.52 $\pm$ 0.60}{8.39}      & \reshl{63.57 $\pm$ 0.49}{2.01}      & \reshl{48.57 $\pm$ 0.98}{6.46}      & \reshl{60.08 $\pm$ 0.36}{7.28}      \\
			\midrule
			Swin Transformer~\cite{liu2021swin}                   & 80.74 $\pm$ 0.00     & 72.22 $\pm$ 0.00     & 88.20 $\pm$ 0.00     & 89.30 $\pm$ 0.00     & 82.6                 \\
			+\Model                      & \reshl{83.26 $\pm$ 0.17}{2.52}      & \reshl{74.42 $\pm$ 0.03}{2.2}       & \reshl{89.90 $\pm$ 0.10}{1.7}       & \reshl{90.69 $\pm$ 0.05}{1.39}      & \reshl{84.57 $\pm$ 0.06}{1.97}      \\
			\midrule
			MobileNet V3~\cite{howard2019searching}                  & 30.6 $\pm$ 0.0       & 26.4 $\pm$ 0.0       & 31.9 $\pm$ 0.0       & 31.2 $\pm$ 0.0       & 30.0               \\
			+\Model                      & \reshl{39.2 $\pm$ 0.3}{8.6}         & \reshl{29.8 $\pm$ 1.3}{3.4}         & \reshl{31.5 $\pm$ 0.1}{-0.4}       & \reshl{38.3 $\pm$ 0.8}{8.2}         & \reshl{34.7 $\pm$ 0.2}{4.7}        \\
			\midrule
			ConvNeXt~\cite{liu2022convnet}         & 67.70 $\pm$ 0.00     & 55.75 $\pm$ 0.00     & 67.51 $\pm$ 0.00     & 53.24 $\pm$ 0.00     & 61.0                 \\
			+\Model                      & \reshl{77.05 $\pm$ 0.18}{9.35}      & \reshl{59.84 $\pm$ 0.77}{4.09}      & \reshl{68.25 $\pm$ 0.20}{0.74}      & \reshl{56.23 $\pm$ 0.48}{2.99}      & \reshl{65.34 $\pm$ 0.27}{4.34}      \\
			\midrule
			ViT-L16~\cite{dosovitskiy2020image}                  & 48.85 $\pm$ 0.00     & 42.48 $\pm$ 0.00     & 51.35 $\pm$ 0.00     & 38.18 $\pm$ 0.00     & 45.2                 \\
			+\Model                      & \reshl{64.57 $\pm$ 1.42}{15.72}     & \reshl{54.77 $\pm$ 0.38}{12.29}     & \reshl{55.06 $\pm$ 0.11}{3.71}      & \reshl{41.06 $\pm$ 0.31}{2.88}      & \reshl{53.86 $\pm$ 0.42}{8.66}      \\
			\midrule
			Mixer-B16~\cite{tolstikhin2021mlp}                  & 26.84 $\pm$ 0.00     & 26.95 $\pm$ 0.00     & 30.07 $\pm$ 0.00     & 22.37 $\pm$ 0.00     & 26.6                 \\
			+\Model                      & \reshl{46.89 $\pm$ 2.11}{20.05}     & \reshl{53.87 $\pm$ 0.18}{26.92}     & \reshl{37.97 $\pm$ 0.99}{7.9}       & \reshl{26.10 $\pm$ 0.17}{3.73}      & \reshl{41.21 $\pm$ 0.48}{14.61}     \\
			\midrule
			Mixer-L16~\cite{tolstikhin2021mlp}                  & 38.18 $\pm$ 0.00     & 26.72 $\pm$ 0.00     & 47.23 $\pm$ 0.00     & 33.97 $\pm$ 0.00     & 36.5                 \\
			+\Model                      & \reshl{57.20 $\pm$ 0.75}{19.02}     & \reshl{36.55 $\pm$ 5.39}{9.83}      & \reshl{49.70 $\pm$ 0.23}{2.47}      & \reshl{36.47 $\pm$ 0.38}{2.5}       & \reshl{44.98 $\pm$ 1.40}{8.48}   \\   
			\bottomrule
		\end{tabular}
	}
	\label{tab:supp-12}
\end{table*}
\end{document}